%% file: tmlr.tex
\documentclass[10pt]{article} % For LaTeX2e
\usepackage[preprint]{tmlr}

\input{math_commands.tex}

\usepackage[utf8]{inputenc} % allow utf-8 input
\usepackage[T1]{fontenc}    % use 8-bit T1 fonts
\usepackage{hyperref}       % hyperlinks
\usepackage{url}            % simple URL typesetting
\usepackage{booktabs}       % professional-quality tables
\usepackage{amsfonts}       % blackboard math symbols
\usepackage{nicefrac}       % compact symbols for 1/2, etc.
\usepackage{microtype}      % microtypography
\usepackage{xcolor}         % colors
\usepackage{enumitem}       % list spacing options (nosep)
\usepackage{wrapfig}        % wraptable / wrapfigure for inline half-width floats
\usepackage{graphicx}       % \includegraphics
\usepackage{caption}        % \captionof for mixing figure+table inside one float
\usepackage{float}          % [H] placement specifier to pin floats in place
\usepackage{amssymb}    % \checkmark
\usepackage{pifont}     % \ding{55} for ✗
\usepackage{titlesec}   % tighten section/subsection vertical spacing

\title{What Accuracy and Gradient Cosine Miss: Evaluating Feedback Alignment via Scale Stability, Reference Validity, and Depth Utility}

\author{\name Yuren Hao \email yurenh2@illinois.edu \\
      \addr Department of Computer Science\\
      University of Illinois at Urbana-Champaign
      \AND
      \name Xiang Wan \email oscarwan@stanford.edu \\
      \addr Department of Computer Science \\
      Stanford University
      \AND
      \name ChengXiang Zhai \email czhai@illinois.edu\\
      \addr Department of Computer Science\\
      University of Illinois at Urbana-Champaign}

\def\month{MM}  % Insert correct month for camera-ready version
\def\year{YYYY} % Insert correct year for camera-ready version
\def\openreview{\url{https://openreview.net/forum?id=XXXX}} % Insert correct link to OpenReview for camera-ready version

\begin{document}

\maketitle

\begin{abstract}
Despite the success of deep learning, training deep networks in biologically plausible and hardware-efficient ways remains an open challenge. Feedback alignment (FA) methods address this by replacing backpropagation's symmetric backward weights with fixed random matrices, but their effectiveness depends critically on whether they can be accurately evaluated. The standard evaluation relies on two quantities: task accuracy and cosine similarity between the method's credit signal and the backpropagation gradient. We show that this reporting pair is insufficient by identifying two independent failure modes, both silent under current reporting: (1) measurement degeneracy, where the BP reference gradient collapses to the numerical floor in terminal-LayerNorm residual architectures, rendering cosine uninterpretable; and (2) aggregation collapse, where the aggregate cosine masks layerwise heterogeneity that concentrates credit at one end of the network. To address these limitations, we propose a diagnostic evaluation protocol based on three checks---scale stability, reference validity, and depth utility---together with per-layer rather than aggregate cosine reporting. Across multiple architectures and methods, the standard reporting pair gives no signal of failure in any audited case, while our protocol identifies all failures with wide calibration margins. The two failure modes are causally independent: a per-block scale penalty alleviates Mode 1 (residual scale explosion driving reference collapse) without affecting Mode 2 (cosine ranking that contradicts every functional metric we measured). Identifying these silent failures prevents researchers from building on non-functional credit assignment and provides actionable guidance for developing FA methods that genuinely train deep layers.
\end{abstract}

\input{sections/introduction}

\input{sections/related-work}

\input{sections/motivating-observations}

\input{sections/failure-modes}

\input{sections/validation}

\input{sections/protocol}

\input{sections/conclusion}

\bibliography{references}
\bibliographystyle{tmlr}

\appendix

\input{appendice/reproducibility}

\input{appendice/per-arch}

\input{appendice/depth-sweep}

\input{appendice/ablations}

\input{appendice/layer0-dominance}

\input{appendice/A}

\input{appendice/sbcb-rescue}

\input{appendice/B}

\input{appendice/depth-ladder}

\input{appendice/threshold-sensitivity}

\input{appendice/pitfalls}

% appendice/C (penalty controls) is intentionally not included --- its
% content now lives in the main text (sections/validation.tex, Sec. 5.2).
%\input{appendice/C}

\end{document}

%% file: math_commands.tex
\usepackage{amsmath,amsfonts,bm}

\def\eqref#1{equation~\ref{#1}}
\def\1{\bm{1}}

\DeclareMathAlphabet{\mathsfit}{\encodingdefault}{\sfdefault}{m}{sl}
\SetMathAlphabet{\mathsfit}{bold}{\encodingdefault}{\sfdefault}{bx}{n}

%% file: sections/introduction.tex
\section{Introduction}

The backpropagation algorithm trains neural networks by propagating error gradients layer by layer from the output back to the input. Despite its effectiveness, BP relies on two mechanisms that are both biologically implausible and computationally constraining: weight symmetry between the forward and backward paths, and sequential layer-by-layer updates that prevent parallelization \citep{lillicrap2016random, hinton2022forwardforward, dellaferrera2022errordriven, whittington2017predictive, belilovsky2018greedy}. Feedback alignment (FA) methods address these constraints by replacing the backward weights with fixed random matrices \citep{lillicrap2016random, bartunov2018assessing, xiao2018biologically, moskovitz2018feedback, crafton2019direct, refinetti2021align}. Direct feedback alignment (DFA) further removes the sequential dependence by projecting the output error directly to each hidden layer through independent random connections \citep{nokland2016direct}, and has been shown to train modern architectures including transformers and graph networks with performance approaching BP \citep{launay2020direct, akrout2019deep}. These properties make the FA family a candidate for both biologically plausible learning and hardware-efficient training \citep{frenkel2021learning}. Yet as these methods scale to deeper residual architectures, a prior question becomes pressing: how should we evaluate whether they actually work?

The standard evidence for a feedback alignment method is two quantities: task accuracy, which measures whether the network learned, and cosine similarity between the method's credit signal and the backpropagation gradient, which measures whether the learning signal points in roughly the right direction \citep{lillicrap2016random, nokland2016direct, launay2020direct, refinetti2021align, bartunov2018assessing}. A method that reaches nontrivial accuracy and reports positive alignment is typically interpreted as having trained the network with useful credit assignment. This pair has been the primary reporting convention across a decade of FA research, and the alignment angle in particular has been explicitly recommended as the basis for training best practices \citep{launay2019principled, lillicrap2016random, refinetti2021align}.

\begin{wrapfigure}{r}{0.5\linewidth}
    \centering
    \vspace{-\baselineskip}
    \includegraphics[width=\linewidth]{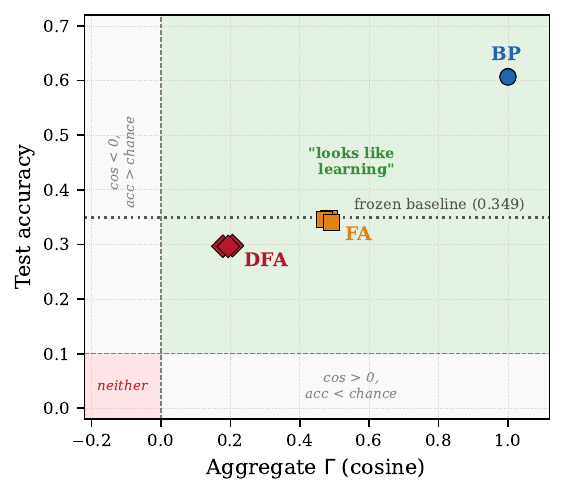}
    \caption{Standard reporting pair on the representative setting ($d{=}512$, $L{=}2$ ResMLP, three seeds): both FA and DFA report non-trivial accuracy and positive aggregate cosine.}
    \label{fig:hero}

    \vspace{4pt}
    \small
    \setlength{\tabcolsep}{4pt}
    \begin{tabular*}{\linewidth}{@{\extracolsep{\fill}}lcccc@{}}
        \toprule
        Method & Test acc & $\Gamma$ & vs Frozen & Verdict \\
        \midrule
        BP  & $0.607$ & $\approx 1.0$ & $+25.7$ pp & \textcolor{green!50!black}{pass} \\
        FA  & $0.345$ & $+0.48$       & $-0.4$ pp  & \textcolor{red!70!black}{fail} \\
        \addlinespace[2pt]
        DFA & $0.297$ & $+0.19$       & $-5.2$ pp  & \textcolor{red!70!black}{fail} \\
        \bottomrule
    \end{tabular*}
    \captionof{table}{Same data as Figure~\ref{fig:hero}; trained blocks of FA and DFA fail to exceed an architecture-matched frozen-random-blocks baseline.}
    \label{tab:audit-headline}
\end{wrapfigure}

The reporting pair gives no signal when a method has not actually trained the network. On a representative residual setting (Figure~\ref{fig:hero}), both FA and DFA report non-trivial accuracy and positive aggregate cosine---the kind of numbers a reader would interpret as evidence of credit reaching the network. Yet neither method exceeds an architecture-matched frozen-blocks baseline whose residual blocks were never trained at all (Table~\ref{tab:audit-headline}): DFA falls $5.2$ percentage points below the baseline, and FA fails to exceed it ($-0.4$ pp, within the baseline's seed variance of $\pm 0.002$). The standard reporting pair shows no sign of this, for two independent reasons. First, the cosine measurement itself can be invalid: in terminal-LayerNorm residual architectures, residual-stream scale explosion compresses the BP reference gradient to the numerical floor through the LayerNorm Jacobian, so the reported cosine is computed against floating-point noise rather than a meaningful direction; and even when the reference is meaningful, an aggregate cosine can appear positive because the per-layer contributions are concentrated at one end of the network. Second, accuracy alone cannot distinguish whether deep blocks help or hurt the network's prediction. Without detecting such failures, researchers risk building on methods whose deep-layer credit is non-functional, misdirecting efforts to scale FA to larger architectures.
%\cheng{Discuss the negative impact of this kind of problems. That is, they don't reveal the detailed problem, but so what? Any negative impact? It would be good to elaborate this so as to strengthen the motivation for studying the failure. E.g., failing to reveal such problems may mislead.... ? }

We demonstrate these failures through a controlled audit of three methods---backpropagation, feedback alignment \citep{lillicrap2016random}, and direct feedback alignment \citep{nokland2016direct}---on residual architectures. FA and DFA share the same local loss and differ only in how credit reaches the deep layers, providing a controlled comparison on which our protocol identifies both failures with wide calibration margins while the standard pair does not. Reliable evaluation is a prerequisite for progress on FA: if the standard metrics cannot distinguish a working method from a non-functional one, positive results in the literature may not reflect genuine advances in credit assignment, and practitioners cannot make informed choices about when FA methods are safe to deploy.
%\cheng{The value of your contribution here isn't clear. To make reviewers appreciate the contribution, you need to first raise some questions (and argue why it's an important question that must be answered) or point out a pain point and then say your contribution addressed that pain point or answered that question. This is perhaps the most important point to convey here; you want to prepare the readers for wanting to see a solution to a problem, and then reveal your solution, which they'd appreciate as a useful contribution. So talking about how important the problem is may be as important as talking about your solution. E.g., if you don't discuss why it's important to study how to cure cancers and the reviewer doesn't know much about what is a cancer or how harmful it is, then they wouldn't appreciate any new research that sheds light on how to cure a cancer. In contrast, if you elaborate what cancer is, and how people currently suffer from it, then even a small progress in curing it would more likely be appreciated. In this particular paper, it's very important to articulate the importance of the problem you addressed since it's not a "usual problem" that many reviewers are familiar with. }

\noindent\textbf{Contributions.}
\begin{itemize}[nosep, topsep=0pt, partopsep=0pt]
    %\cheng{again try to say why we need to identify such failure modes; what if we don't bother to identify them? At least say something like "To improve or enhance FA training, we identify .." or "To determine when we should use FA, we identify ..." (I'm not sure any of these make sense, but you should clearly articulate the pain point and also discuss how exactly your contribution would address the pain point.}
    \item To enable reliable diagnosis of whether FA methods genuinely train deep layers, we identify two independent failure modes of the standard FA evaluation pair (accuracy + aggregate cosine) on deep residual architectures: \emph{measurement degeneracy}, where the BP reference gradient collapses to the numerical floor and makes cosine uninterpretable, and \emph{low intrinsic credit-direction quality}, where deep-layer credit is essentially unaligned with BP even when the reference is meaningful.
    \item We provide a three-diagnostic evaluation protocol that detects both modes. On a representative setting where both FA and DFA report non-trivial accuracy and positive cosine alignment, the standard pair gives no signal that either method's deep blocks are unused; the protocol identifies both as failures with wide calibration margins.
    \item We show that cosine alignment to the BP gradient, even when measured validly, does not predict depth utility: under matched conditions, three independent functional metrics rank the audited methods in an order that cosine contradicts.
    \item We validate the protocol across multiple architectures, methods, and experimental settings, and release a reference implementation that enables practitioners to apply the protocol to new architectures and methods.
    %\cheng{I'm not sure what kind of contribution this is. Why is this reference implementation useful? Say something like "release a ref. implementation which enables ...". }
\end{itemize}

%% file: sections/related-work.tex
\section{Related Work}
\label{sec:related-work}

\textbf{Feedback alignment family.}
\citet{lillicrap2016random} introduced feedback alignment by showing that fixed random backward weights can replace the transpose of the forward weights and still support learning, provided the forward weights learn to align with the random feedback. The theoretical basis for this tolerance was anticipated by \citet{liao2015important}, who systematically studied the importance of weight symmetry in backpropagation and found that several asymmetric variants still learn. \citet{nokland2016direct} removed the sequential dependence entirely with direct feedback alignment (DFA), projecting the output error directly to each hidden layer. Subsequent work extended FA and DFA along several axes: \citet{bartunov2018assessing} assessed the scalability of FA to deeper architectures and harder tasks, finding degradation relative to BP at scale; \citet{moskovitz2018feedback} analyzed feedback alignment in deep convolutional networks; \citet{xiao2018biologically} showed that biologically plausible algorithms can scale to large datasets; \citet{crafton2019direct} introduced sparse feedback connections for local learning; and \citet{akrout2019deep} proposed weight-mirror mechanisms that learn symmetric backward weights without explicit weight transport. \citet{launay2020direct} demonstrated that DFA can train modern architectures including transformers with performance approaching BP, and \citet{wang2024streamlined} recently extended this line to optical training of large-scale architectures. \citet{robertson2023implicit} provided a theoretical lens on FA through implicit regularization, showing that feedback alignment induces qualitatively different implicit biases from BP. Most recently, \citet{caillon2026scaling} introduced Jacobian alignment guarantees for scaling DFA to deeper architectures, and \citet{boeshertz2026overcoming} showed how to overcome the rank collapse that limits aligned feedback---both advancing the depth-scaling frontier that our audit quantifies from the evaluation side. On the hardware side, fixed random feedback removes weight transport and enables feedforward-only training on neuromorphic and energy-efficient chips \citep{frenkel2021learning, park2020neuromorphic}. This line stays active across biological plausibility, depth scaling, and dedicated hardware, yet the evaluation methodology shared across it has gone unexamined; we audit that methodology rather than propose a new FA variant.

\textbf{Biological plausibility alternatives.}
Feedback alignment is one of several families of biologically plausible learning algorithms. Target propagation \citep{lee2014difference} replaces backward error gradients with layer-local target activations, avoiding weight transport entirely; \citet{meulemans2020theoretical} later provided a theoretical framework unifying several TP variants and analyzing their convergence properties. \citet{whittington2017predictive} showed that predictive coding networks with local Hebbian plasticity can approximate backpropagation. \citet{scellier2017equilibrium} proposed equilibrium propagation, which bridges energy-based models and backpropagation using only local computations at equilibrium; \citet{liu2025toward} recently improved its practicality with feedback regulation and residual connections. \citet{belilovsky2018greedy} demonstrated that greedy layerwise training can scale to ImageNet without end-to-end backpropagation. More recently, \citet{hinton2022forwardforward} introduced the forward-forward algorithm, replacing the forward and backward passes with two forward passes operating on positive and negative data, and \citet{dellaferrera2022errordriven} proposed error-driven input modulation, solving credit assignment without a backward pass. These methods share with FA the goal of relaxing BP's constraints, but differ in which constraints they target (weight symmetry, global error signals, or backward locking). Our diagnostic protocol is designed for the FA family specifically, where cosine alignment to the BP gradient is the standard evaluation metric; extending it to other families would require analogous reference quantities.

\textbf{Evaluation methodology for FA.}
The two-quantity reporting convention---task accuracy and cosine alignment to the BP gradient---has been the primary evaluation standard since \citet{lillicrap2016random} introduced the alignment angle as evidence that forward weights rotate toward the random backward matrices. \citet{refinetti2021align} provided a theoretical account of the alignment dynamics, showing a two-phase process (align, then memorise) and using per-layer cosine trajectories as the primary diagnostic. \citet{launay2019principled} explicitly recommended the alignment angle as the basis for training best practices in DFA, deriving principled hyperparameter choices from cosine measurements. Per-layer cosine reporting appears in several mechanism studies \citep{lillicrap2016random, refinetti2021align} but has not been adopted as a standard reporting requirement. \citet{bartunov2018assessing} came closest to a systematic audit by evaluating multiple biologically plausible methods on common benchmarks and noting FA's degradation at scale, but their analysis focused on task accuracy rather than diagnosing why the credit signal fails; no subsequent work has made the reporting protocol itself the object of audit. Our contribution is orthogonal to these methodological advances: we show that cosine alignment---whether reported in aggregate or per-layer---can be uninformative due to measurement degeneracy (Mode~1) and does not predict depth utility even when valid (Section~\ref{subsec:cosine-insufficient}), motivating a protocol that adds scale stability and frozen-blocks checks alongside cosine reporting.

%% file: sections/motivating-observations.tex
\section{Motivating Observations}
\label{sec:motivating}
%\cheng{Need to motivate in a way that's understandable and appreciated by people who aren't working on training neural networks themselves. Due to shortage of reviewers, papers are often reviewed by people who aren't necessarily working on the exact topic. So it would be useful to start with some practical challenge or pain point, and gradually connect it with the low-level details. As is, you just talk about the process you followed, i.e., what you did, but no (much) explanation of why you want to do this and what new knowledge we can expect to discovery by doing this. Make sure to always explain the motivation clearly before talking about how exactly you did X.  }
To test whether the standard reporting pair can silently pass methods that have not trained the network, we design a controlled audit: three credit rules sharing identical architecture and training, differing only in how credit reaches each layer. We fix a common training recipe across all primary-audit experiments: pre-LayerNorm \citep{ba2016layernorm, xiong2020prelayernorm} ResMLPs \citep{touvron2021resmlp} on CIFAR-10 \citep{krizhevsky2009cifar}, trained for 100 epochs with AdamW \citep{loshchilov2019adamw} (learning rate $10^{-3}$, weight decay $0.01$, cosine schedule, batch size 128) across three seeds (full architecture and training details in Appendix~\ref{app:reproducibility}). Architecture depth $L$ and width $d$ are specified per experiment. Three methods are trained on this identical setup, differing only in how each layer's credit signal $a_l$ is computed.

\setlength{\intextsep}{0pt}
\begin{wrapfigure}{r}{0.42\linewidth}
    \centering
    \vspace{-\baselineskip}
    \includegraphics[width=\linewidth]{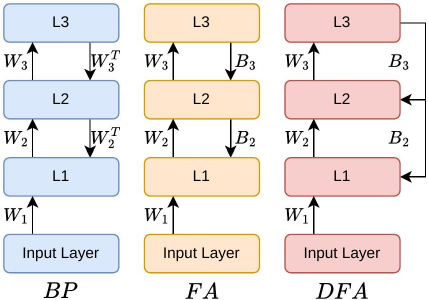}
    \caption{Backward architectures}
    \label{fig:credit-paths}
\end{wrapfigure}
\setlength{\intextsep}{3pt plus 1pt minus 1pt}

\textbf{Backpropagation (BP)} computes the exact gradient via the chain rule: $a_l = \partial L / \partial h_l$.
%\cheng{this part reads like a sequence of definitions; try to make a story by increasing the cohesion. E.g., below we first give definitions of BP, FA and DFA and then ..... }
The two FA variants replace BP's exact backward pass with progressively simpler credit paths (Figure~\ref{fig:credit-paths}):

\textbf{Feedback Alignment (FA)} \citep{lillicrap2016random} propagates credit sequentially through fixed random matrices: starting from the exact output-layer gradient $\partial L / \partial h_L$, each layer receives $a_l = B_l \, a_{l+1}$, where $B_l \in \mathbb{R}^{d \times d}$ is fixed at initialization.

\textbf{Direct Feedback Alignment (DFA)} \citep{nokland2016direct} projects the output error directly to each layer, bypassing all intermediate layers: $a_l = B_l^\top e_T$, where $B_l \in \mathbb{R}^{C \times d}$ is a fixed random matrix and $e_T = \hat{y} - y$ is the output error.

FA and DFA share the same local loss: each block $f_l$ is updated by reducing $-\langle f_l(h_l), a_l \rangle$. Neither loss contains a penalty on $\|f_l(h_l)\|$. The only difference between the two methods is how $a_l$ is computed: FA preserves sequential structure in the credit path, DFA does not. A frozen-blocks baseline trains only the embedding, LayerNorm, and classification head while holding all residual blocks fixed at their random initialization; it reaches $0.349 \pm 0.002$ across the same three seeds.

\begin{figure}[h]
    \centering
    \includegraphics[width=0.9\linewidth]{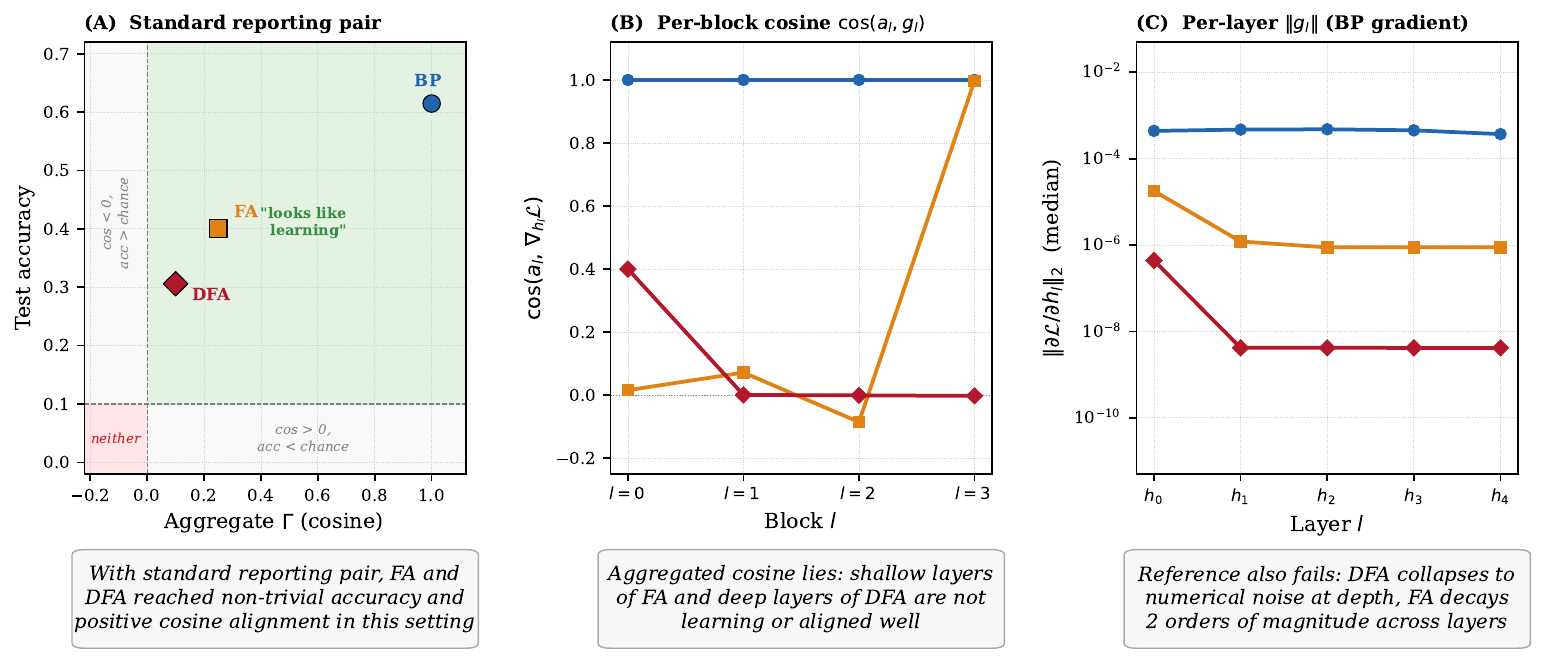}
    \caption{Per-layer state on the standard primary-audit setting (ResMLP $d=256$, $L=4$, CIFAR-10). (A) Standard reporting pair: all three methods report non-trivial accuracy and positive aggregate cosine. (B) Per-block cosine reveals that FA's shallow blocks and DFA's deep blocks contribute no aligned credit. (C) Per-layer BP gradient norm decays across FA's depth and collapses to numerical noise in DFA, leaving cosine measurements at depth without a meaningful reference.}
    \label{fig:overview}
\end{figure}

%\cheng{here it doesn't sound like you are discovering something new or significant. You give the impression that these are just ordinary results that you observed in some experiments. No research question implied here.
%If nobody studied this failure, your research question should be something like "are there "hidden"/"silent" failures"?  Then say, knowing the answer to this question is important because .. and no existing work has given any answer. We study this question and use xxx experiments to analyze the potential hidden failures. We found that there are two .... and knowing that there are these two failures is beneficial because .... This kind of story line should be told explicitly and clearly to help readers appreciate the contributions.  }
The failure shown in Figure~\ref{fig:hero} has two independent sources internal to the standard reporting pair. To isolate them, we turn to the per-layer state of a related setting at $d=256, L=4$ (Figure~\ref{fig:overview}). Two independent inconsistencies emerge from the per-layer data, each one severing the link between the standard reporting pair and the question it is meant to answer.

\textbf{The aggregate cosine masks opposite per-layer patterns.} On this setting, FA reports an aggregate cosine of $\Gamma \approx +0.23$ and DFA reports $\Gamma \approx +0.10$; both look like partial alignment to the BP gradient. Per-block cosine reveals that these aggregates are produced by inverse distributions (Figure~\ref{fig:overview}B). FA's shallowest block contributes near-zero cosine ($\approx +0.01$) while its deepest block contributes $+0.95$; the aggregate is carried almost entirely by the deep end of the network. DFA shows the opposite shape: its shallowest block contributes $+0.39$ while its deepest block contributes $\approx 0$; the aggregate is carried almost entirely by the embedding. Two networks whose credit reaches opposite ends of the depth axis produce aggregate cosines that read as the same kind of evidence. Aggregation collapses this distinction, so the aggregate $\Gamma$ cannot in principle answer whether credit reaches the deep layers. The same layer-0-dominated profile holds across depths $L\in\{2,4,6,8,12\}$ (Appendix~\ref{app:depth-sweep}).

\textbf{The reference gradient can collapse below the cosine clamp.} The BP gradient norm $\|g_l\|$ along depth tells the second story (Figure~\ref{fig:overview}C). BP holds $\|g_l\|$ within an order of magnitude of $\sim 4\times 10^{-4}$ across all layers, providing a reference of consistent scale. FA decays from $\sim 2\times 10^{-5}$ at the embedding to $\sim 9\times 10^{-7}$ at the deepest layer, but every layer remains above PyTorch's default cosine-similarity denominator floor of $\varepsilon = 10^{-8}$. DFA collapses by three orders of magnitude after the embedding---from $\sim 4\times 10^{-7}$ at $h_0$ to $\sim 4\times 10^{-9}$ at $h_1$ and below---and remains beneath $\varepsilon$ for every deeper layer. When the denominator is supplied by the clamp rather than the reference vector, $\cos(a_l, g_l)$ is no longer comparing the credit direction to a meaningful BP direction; it is comparing it to a normalized floating-point residual.

\textbf{The two failures are independent.} On this setting, FA fails the first check (its aggregate is dominated by a single end of the network) but passes the second (every reference norm remains measurable); DFA fails both. The aggregate-dominance failure does not require reference collapse, and reference collapse does not require any particular layerwise distribution of cosine. Each one severs the standard reporting pair from its intended meaning on its own. Neither method universally avoids either mode---Section~\ref{sec:validation} shows that both modes can be triggered by either method in other settings, and that interventions on one mode do not resolve the other.\looseness=-1

%\cheng{I don't see what this section motivates. It shows the existence of these failures and I can see they can motivate a question like " So, now that we know there failures exist, how can we detect *all* such failures? This question hasn't been studied and we propose a method for doing this....In any case, to summarize this section, you should raise a research question, which you then answer/address in the next section.  }
These observations raise a concrete question: can the standard reporting pair be supplemented with diagnostics that detect both failure modes and determine whether trained depth is genuinely useful? We address this in the following sections by first characterizing each failure mode as a general mechanism, then proposing a protocol that detects both.

%% file: sections/failure-modes.tex
\section{Failure Modes}
\label{sec:failure-modes}
%\cheng{It would help if you make sure to always give a short overview of the section at the very beginning, ideally connecting it with the previous section to make a coherent story. Something like: In the previous section, we see there are two failures. In this section, we go deeper or systematically address them or .show that these are two *general* failures that always exist ... At the end of a section, it's also often useful to summarize a section with a takeaway which usually motivates the next section. If you pay attention to the beginning and end of each section to ensure the high-level cohesion, it would make it much easier for readers to follow your paper/story. }
The previous section showed that the standard reporting pair fails through two independent inconsistencies on a specific setting. We now characterize each as a general failure mode with a distinct causal mechanism, explaining when and why each arises.

\begin{figure}[H]
    \centering
    \includegraphics[width=\linewidth]{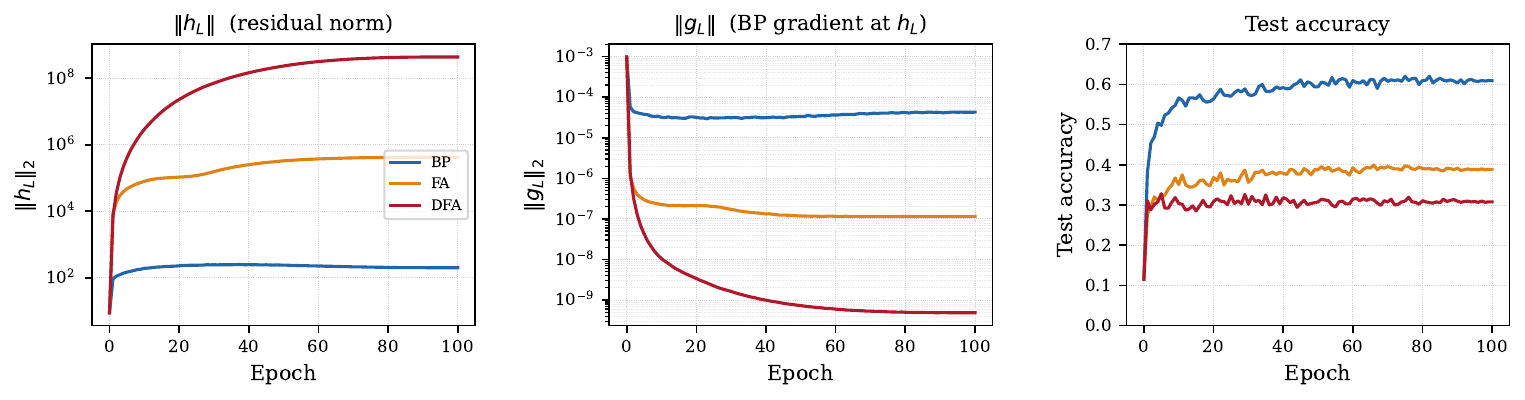}
    \caption{Temporal evolution of $\|h_L\|$, $\|g_L\|$, and test accuracy on ResMLP $d=256$, $L=4$, terminal LN.}
    \label{fig:temporal-resmlp}
\end{figure}

\subsection{Mode 1: Residual scale explosion drives reference collapse}
\label{subsec:mode1}

In Mode 1, the BP reference gradient at deep layers is compressed below the numerical floor used by the cosine implementation, so the cosine reported by the standard pair is computed against a normalized floating-point residual rather than against the BP direction. The mechanism is a chain of three steps connecting the local loss form to this measurement collapse.

The local loss $-\langle f_l(h_l), a_l \rangle$ is unbounded in $\|f_l(h_l)\|$: any direction in which the block output grows along $a_l$ reduces the loss further. In a pre-LN residual block $h_{l+1} = h_l + f_l(h_l)$, growth in $\|f_l(h_l)\|$ accumulates directly into the residual stream, so $\|h_L\|$ increases along training. The terminal LayerNorm normalizes $h_L$ by its per-coordinate standard deviation $\sigma \propto \|h_L\|/\sqrt{d}$, so its Jacobian's nonzero singular values scale as $\sqrt{d}/\|h_L\|$ \citep{xiong2020prelayernorm}. The deepest BP reference gradient $\|g_L\| = \|\partial L/\partial h_L\|$ inherits this factor and shrinks as $1/\|h_L\|$, so the product $\|h_L\|\,\|g_L\|$ stays roughly constant as the residual stream explodes.\looseness=-1

Figure~\ref{fig:temporal-resmlp} shows this chain in the DFA trajectory on the primary-audit setting: $\|h_L\|$ grows from $\sim 10^2$ at initialization to $\sim 10^8$ by epoch 100, while $\|g_L\|$ collapses from $\sim 10^{-3}$ to $\sim 10^{-9}$ over the same window---four orders of magnitude in each direction, in tandem. By epoch 20, $\|g_L\|$ is already below PyTorch's default cosine-similarity denominator floor of $\varepsilon = 10^{-8}$, and remains there for the rest of training.\looseness=-1

FA shares DFA's local loss form but exhibits a markedly attenuated chain on the same architecture (Figure~\ref{fig:temporal-resmlp}): $\|h_L\|$ grows to $\sim 10^5$ rather than $\sim 10^8$, and $\|g_L\|$ stabilizes at $\sim 10^{-7}$, within the regime where cosine is computed against a meaningful reference. The local loss form is therefore not sufficient for Mode 1 to become a measurement failure; the credit propagation rule determines whether the chain reaches the cosine clamp. FA is not generally safe from Mode 1---the same mechanism produces full reference collapse on terminal-LN architectures with different geometry (Section~\ref{sec:validation}).\looseness=-1

\subsection{Mode 2: Aggregation collapses layerwise heterogeneity}
\label{subsec:mode2}

In Mode 2, the per-layer cosine values along depth concentrate at one end of the network, and the aggregate cosine averages over this concentration; $\Gamma$ no longer indicates whether credit reaches the deep layers. Section~\ref{sec:motivating} showed two such concentrations: FA's cosine is carried by the deep end, DFA's by the shallow end, yet both produce similar-looking aggregates.

The two concentrations follow from the credit propagation rule. DFA's $a_l = B_l^\top e_T$ is a fixed random projection of the output error and contains no information about the forward state at layer $l$. The embedding sits one block away from the output and receives credit whose correlation with the BP gradient is preserved by the projection's geometry, but every deeper block receives a signal generated independently of its forward state, and the cosine reflects that independence---near zero in expectation. FA's $a_l = B_l \, a_{l+1}$ inherits the deepest block's exact gradient $\partial L/\partial h_L$ and degrades upstream by an additional random matrix product per layer. The two rules produce alignment patterns that are inverse but symmetric in their effect on the aggregate.\looseness=-1

Per-layer cosine reporting \citep{lillicrap2016random, refinetti2021align} addresses this specific aggregation failure but does not recover the standard pair: it carries no information about whether each layer's reference gradient is numerically valid (Mode 1), and it does not indicate whether the trained depth as a whole contributes to the network's prediction (Section~\ref{sec:protocol}). Section~\ref{subsec:penalty} establishes that Mode 1 and Mode 2 are causally independent failure modes, both observationally and under a penalty intervention that alleviates Mode 1 without affecting Mode 2.\looseness=-1

%% file: sections/validation.tex
\section{Validation}
\label{sec:validation}
%\cheng{Again, please add a small paragraph at the beginning to talk about what kind of validation you want to do, and why (if it's not obvious). That is, framing a research question first before elaborating what you have done. The goal is not to just talk about what you did or what you observed, but to raise a research question, address it, discuss how well you address it, and then motivate the next question, discuss how you address it, and how well you address it, etc. Think about the paper as a sequence of connected questions and their answers. As is, there are a lot of answers, but there are few or no questions being raised. Experts on the topic may infer the questions being addressed, but ordinary reviewers wouldn't be able to do it. So you need to articulate the questions or challenges you addressed explicitly, considering the reviewers assigned are often outside the core topic area (note that they are very different from a reader who finds your paper interesting and decides to read it! Such readers are already motivated to read it. Reviewers were "forced" to read it sometimes, so they might not be naturally motivated.  }
The failure modes documented on a single architecture raise three questions: (1)~do they arise on other architectures, or are they specific to ResMLP? (2)~are they causally independent, as the observational evidence suggests? (3)~does cosine alignment, even when measured validly, suffice to predict whether depth is useful? We address each in turn.

\subsection{Scope of Mode 1 across architectures}
\label{subsec:mode1-scope}

\begin{figure}[h]
    \centering
    \includegraphics[width=\linewidth]{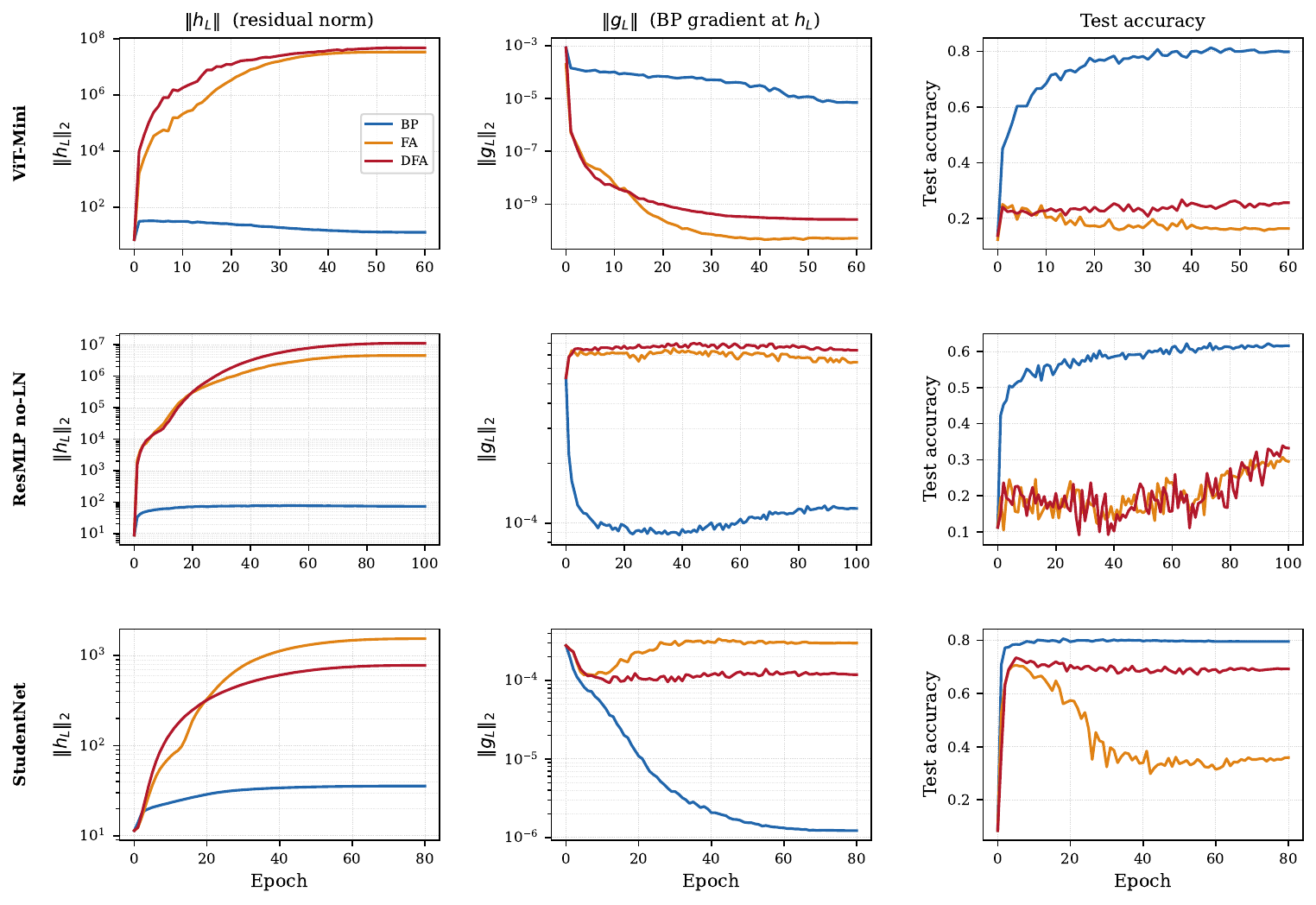}
    \caption{Temporal evolution of $\|h_L\|$, $\|g_L\|$, and test accuracy across three architectures: ViT-Mini (terminal LN), ResMLP no-LN, and StudentNet (no LN). $\|g_L\|$ y-axis ranges differ across rows.}
    \label{fig:crossarch}
\end{figure}

The two failure modes documented in Section~\ref{sec:failure-modes} on ResMLP have different cross-architecture scopes. We examine three additional settings that vary terminal LayerNorm presence while holding the other components of the architecture roughly fixed (Figure~\ref{fig:crossarch}): a vision transformer ViT-Mini \citep{dosovitskiy2021vit} ($d=128, L=4$, cls token + terminal LN), the same ResMLP $d=256, L=4$ used in the primary audit but with the terminal LayerNorm removed, and StudentNet ($d=128, L=4$, no LN). Across these three settings, Mode 1a (residual scale growth under FA and DFA) appears in all three; Mode 1b (reference gradient collapse) appears only in the architecture with terminal LayerNorm.

On ViT-Mini, both FA and DFA reproduce the full Mode 1 chain seen on ResMLP: $\|h_L\|$ grows to $\sim 10^7$ and $\|g_L\|$ collapses to $\sim 10^{-9}$ within 30 epochs, well below the regime where cosine similarity computes against a meaningful reference. The transformer architecture provides no protection from Mode 1 once terminal LN is in place. On ResMLP with terminal LN removed and on StudentNet (no LN), the residual stream still grows under FA and DFA---$\|h_L\|$ reaches $\sim 10^7$ on no-LN ResMLP and $\sim 10^3$ on StudentNet---but $\|g_L\|$ stabilizes at or above $\sim 10^{-4}$ on both, in the regime where cosine remains a valid measurement. The unbounded growth incentive of the local loss is architecture-agnostic, but its translation into measurement collapse requires terminal LayerNorm to compress the BP reference gradient. The growth also survives removing the residual skip and replacing the labels with random targets (Appendix~\ref{app:ablations}).\looseness=-1

\begin{wraptable}{r}{0.5\linewidth}
    \centering
    \vspace{-\baselineskip}
    \small
    \setlength{\tabcolsep}{4pt}
    \begin{tabular*}{\linewidth}{@{\extracolsep{\fill}}lccc@{}}
        \toprule
        Architecture & LN & Scale stable & Reference valid \\
        \midrule
        ResMLP     & yes & \textcolor{red!70!black}{\ding{55}} / \textcolor{red!70!black}{\ding{55}} & \textcolor{green!50!black}{\checkmark} / \textcolor{red!70!black}{\ding{55}} \\
        ResMLP, no LN & no & \textcolor{red!70!black}{\ding{55}} / \textcolor{red!70!black}{\ding{55}} & \textcolor{green!50!black}{\checkmark} / \textcolor{green!50!black}{\checkmark} \\
        ViT-Mini                   & yes & \textcolor{red!70!black}{\ding{55}} / \textcolor{red!70!black}{\ding{55}} & \textcolor{red!70!black}{\ding{55}} / \textcolor{red!70!black}{\ding{55}} \\
        StudentNet                 & no & \textcolor{red!70!black}{\ding{55}} / \textcolor{red!70!black}{\ding{55}} & \textcolor{green!50!black}{\checkmark} / \textcolor{green!50!black}{\checkmark} \\
        \bottomrule
    \end{tabular*}
    \caption{Mode 1 verdict per architecture (FA/DFA).}
    \label{tab:arch-summary}
\end{wraptable}

Table~\ref{tab:arch-summary} (\textcolor{green!50!black}{\checkmark}=passes, \textcolor{red!70!black}{\ding{55}}=fails, FA/DFA) shows Mode 1a failing universally---the unbounded growth incentive is architecture-agnostic---while Mode 1b tracks terminal LN presence. The same-backbone control isolates this: removing terminal LN from the primary-audit ResMLP flips the Mode 1b verdict for both methods without changing Mode 1a. Appendix~\ref{app:per-arch} reports the per-architecture diagnostic values. The depth-utility consequence on ViT-Mini is severe---the frozen-blocks baseline reaches $0.570 \pm 0.003$ while FA and DFA reach $0.163$ and $0.256$, trained blocks underperforming random blocks by more than $30$ percentage points.

\subsection{Penalty intervention separates the two modes}
\label{subsec:penalty}

The two failure modes can be intervened on independently. Adding a per-block scale penalty $\lambda \cdot \|f_l(h_l)\|^2$ to the local loss on each residual block---without modifying the credit rule, the optimizer, or any other component---suppresses Mode 1 in a dose-response manner on methods where Mode 1 is severe; at the dose where Mode 1 is fully alleviated but Mode 2 is not yet engaged, the deep-layer cosine remains at the vanilla value, separating the two modes causally.

\begin{figure}[H]
    \centering
    \includegraphics[width=\linewidth]{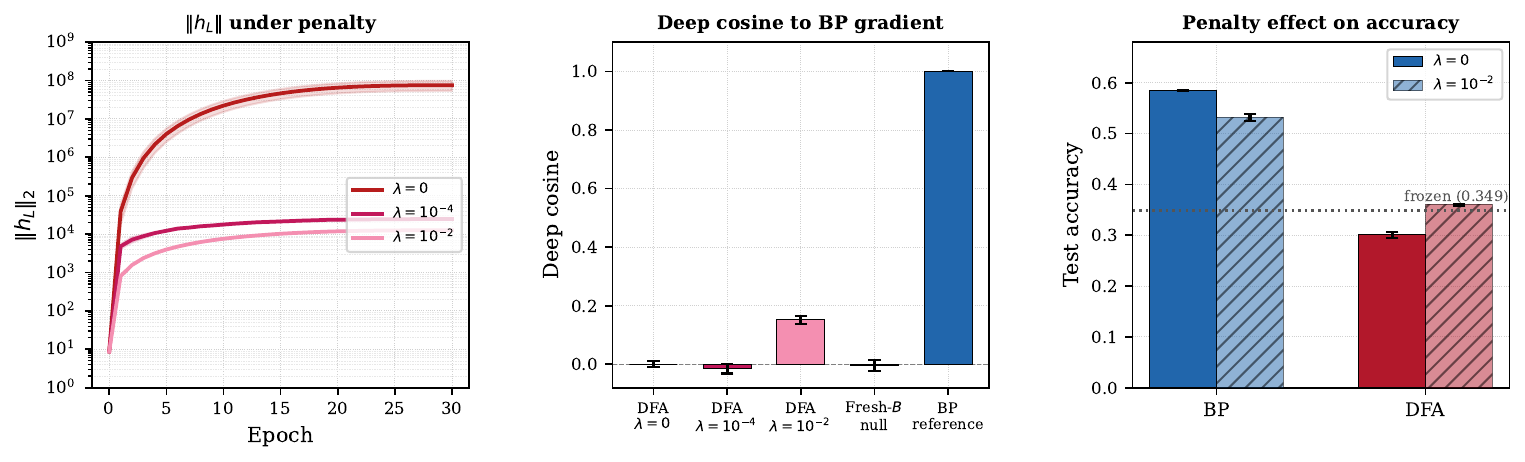}
    \caption{Penalty intervention on DFA, ResMLP $d=256, L=4$, three seeds, 30 epochs. \textbf{Left:} $\|h_L\|$ under $\lambda \in \{0, 10^{-4}, 10^{-2}\}$. \textbf{Middle:} deep cosine across conditions. \textbf{Right:} test accuracy against frozen-blocks baseline.}
    \label{fig:penalty}
\end{figure}

We sweep $\lambda \in \{0, 10^{-4}, 10^{-2}\}$ on DFA, ResMLP $d=256, L=4$, three seeds, 30 epochs (Figure~\ref{fig:penalty}, left and middle panels). Vanilla DFA reaches $\|h_L\| \sim 4 \times 10^8$ and $\|g_L\| \sim 5 \times 10^{-10}$, with deep cosine at the vanilla zero. At $\lambda = 10^{-4}$, the residual stream is contained ($\|h_L\| \sim 2.3 \times 10^4$) and $\|g_L\| \sim 5 \times 10^{-6}$ recovers well above the clamp---Mode 1 is alleviated---while deep cosine remains at $-0.016 \pm 0.016$, indistinguishable from zero. At $\lambda = 10^{-2}$, $\|h_L\|$ is further contained and deep cosine rises to $+0.152 \pm 0.013$, indicating partial recovery of Mode 2. The two modes respond at different intervention strengths, and Mode 2 does not follow Mode 1's recovery passively. FA on ResMLP exhibits only mild Mode 1 and produces no dose response (Appendix~\ref{app:fa-penalty}).

A potential alternative explanation for the deep-cosine recovery on DFA at $\lambda = 10^{-2}$ is that the penalty itself shifts cosine measurement statistics independent of any change in credit quality. We rule this out with a BP control: BP under the same penalty schedule reaches $0.585 \pm 0.001$ accuracy at $\lambda = 0$ and $0.532 \pm 0.007$ at $\lambda = 10^{-2}$, a capacity cost of $5.3$ percentage points (Figure~\ref{fig:penalty}, right panel). BP's deep cosine remains $\approx 1.0$ throughout: the penalty does not artificially lift cosine on a method whose credit is already exact. The penalty is not a free intervention---it costs accuracy on every method we tested---but BP+penalty still exceeds the frozen-blocks baseline by $18.3$ percentage points, so the penalty does not turn working methods into broken ones.

A second alternative is that the recovered deep cosine reflects trivial adaptation to the fixed feedback matrices $B_l$ rather than an improvement in credit quality. We rule this out with a fresh-$B$ null calibration: holding the trained DFA network at the $\lambda = 10^{-2}$ checkpoint (seed 42) fixed and replacing each $B_l$ with a freshly drawn random matrix gives deep cosine $-0.005 \pm 0.018$ over 20 draws, against the trained-$B$ value of $+0.166$ on the same checkpoint---a $9.3\sigma$ separation. The recovered cosine is well above the noise floor and is specific to the matrices used during training.

Two observational checks corroborate the intervention. Across methods on ResMLP $d=256, L=4$, FA's $\|g_l\|$ remains above the cosine clamp at every layer (Figure~\ref{fig:temporal-resmlp}, middle panel), so its standard pair fails only through Mode 2; DFA on the same architecture fails both. Along training time, DFA's deep-layer cosine is already $-0.008 \pm 0.013$ at epoch 1 while $\|g_L\| \sim 10^{-7}$ remains above $\varepsilon$: Mode 2 is present before Mode 1 develops; Appendix~\ref{app:layer0} gives the per-seed per-layer cosines. Together with the penalty intervention, this establishes Mode 1 and Mode 2 as causally independent failure modes the protocol must diagnose separately.\looseness=-1

\subsection{Cosine cannot predict depth utility}
\label{subsec:cosine-insufficient}

%\cheng{the following motivating paragraph is excellent! The paper would benefit from having many more such paragraphs. Still, one might also wonder what it's useful to know cosine can't predict depth utility? Is it just an argument that cosine is not enough? If so, you should explicitly say so and then say to address this issue, we propose additional measures including ..... In other words, making everything very explicit using plain English to more clearly communicate the problems you addressed, your solutions, and the potential impact of your solutions.}
Mode 1 and Mode 2 together explain why the standard reporting pair fails on settings where the measurement is invalid or the aggregate masks layerwise heterogeneity. A natural follow-up is whether cosine alignment, when measured validly and reported per-layer, suffices on its own---that is, whether the depth-utility check (the frozen-blocks comparison) is redundant once Mode 1 and Mode 2 are addressed. We show it is not: even on settings where Mode 1 has been alleviated and the cosine measurement is valid, deep-layer cosine to the BP gradient does not predict whether depth is being used.

To establish this, we introduce two diagnostic probes designed to vary credit quality in the cosine space: 

\textbf{State Bridge (SB)} learns a state predictor $G_\psi(h_l, t_l)$ that estimates the deepest hidden state $h_L$ from the current layer's state; the credit signal is the gradient of a cross-entropy loss computed by passing $G_\psi(h_l, t_l)$ through the classification head, $a_l = \nabla_{h_l} \mathrm{CE}(\mathrm{head}(G_\psi(h_l, t_l)), y)$. 

\textbf{Credit Bridge (CB)} learns a value network $V_\phi(h_l, t_l)$ that directly estimates a scalar value at the current layer; the credit signal is its input gradient, $a_l = \nabla_{h_l} V_\phi(h_l, t_l)$. 

SB and CB are not proposed as competitive FA methods; they exist to produce credit signals whose deep-layer cosine to the BP gradient differs systematically from DFA's, so that we can ask whether these cosine differences correspond to functional differences in credit quality. Both probes train under the same per-block penalty $\lambda = 10^{-2}$ used in Section~\ref{subsec:penalty} to alleviate Mode 1, the predictor and value network themselves are unpenalized.\looseness=-1

\begin{table}[h]
    \centering
    \small
    \setlength{\tabcolsep}{6pt}
    \caption{Four methods under matched penalty rescue ($\lambda = 10^{-2}$, ResMLP $d=256, L=4$, three seeds, 30 epochs). Three functional metrics agree on SB $\gg$ rest; deep cosine ranks the methods in a different order.}
    \label{tab:dissociation}
    \begin{tabular*}{\linewidth}{@{\extracolsep{\fill}}lcccc@{}}
        \toprule
        Method & Test acc & Deep cosine & Nudging ($\eta=0.01$) & Train loss $\Delta$ \\
        \midrule
        SB + pen  & $\mathbf{0.453 \pm 0.003}$ & $+0.322 \pm 0.008$ & $\mathbf{-1.93 \times 10^{-3}}$ & $\mathbf{-0.447}$ \\
        FA + pen  & $0.369 \pm 0.003$ & $+0.423 \pm 0.006$ & $-2.09 \times 10^{-4}$ & $-0.128$ \\
        CB + pen  & $0.360 \pm 0.004$ & $\mathbf{+0.679 \pm 0.010}$ & $-4.26 \times 10^{-4}$ & $-0.121$ \\
        DFA + pen & $0.360 \pm 0.002$ & $+0.152 \pm 0.013$ & $-4.98 \times 10^{-5}$ & $-0.095$ \\
        \bottomrule
    \end{tabular*}
\end{table}

Table~\ref{tab:dissociation} reports four methods under matched penalty rescue; Appendix~\ref{app:sbcb-rescue} adds the per-seed State Bridge and Credit Bridge values with residual-scale columns. The three functional metrics---test accuracy, the test-loss change from one $\eta{=}0.01$ step along each layer's credit direction (nudging), and the per-run training-loss decrease---agree on a single ranking: SB $\gg$ FA $\approx$ CB $\approx$ DFA, with SB reaching $0.453$ accuracy ($9$ pp above the others) and $4$--$40\times$ larger nudging effects. Deep cosine ranks them differently: CB $>$ FA $>$ SB $>$ DFA. CB has the highest cosine ($+0.679$) but matches DFA's accuracy; SB has lower cosine than FA ($+0.322$ vs $+0.423$) but $9$ pp higher accuracy. Cosine measures angular agreement with BP at a single point, which does not predict whether updates in that direction compose into useful depth training. The depth-utility check therefore cannot be replaced by a per-layer cosine check, even when Mode 1 is alleviated: whether trained depth contributes to prediction is a question about functional behavior, not angular agreement, requiring the frozen-blocks baseline as its own measurement.

%% file: sections/protocol.tex
\section{Recommended evaluation protocol}
\label{sec:protocol}

%\cheng{Once again, you need an introductory paragraph here to connect it with the previous section and give an overview of what the reader expects to see in this section.  It would be especially useful to mention here why it's so important to develop diagnostic checks. What if we don't do this? Would there be any pain? Or if we could do this, what would be the benefit? Having this kind of discussion would strengthen the motivation and help readers appreciate the value of developing 3 diagnostic checks. }
Without diagnostic checks, the two failure modes identified above remain invisible under the standard reporting pair, leaving researchers unable to distinguish methods that train the network from methods that do not. The audit and the failure-mode analysis converge on three checks that together address how the standard reporting pair fails. We state them as a protocol that an evaluator can run on any FA training run; the design rationale follows the structure of Sections~\ref{sec:failure-modes} and \ref{sec:validation}.\looseness=-1

\subsection{Three diagnostic checks}
\label{subsec:diagnostics}

\textbf{Diagnostic 1 (Scale stability).} Compute the maximum per-block residual growth $\rho = \max_l \|h_{l+1}\|/\|h_l\|$ at the end of training. Flag the run if $\rho > 50$. This detects the residual-stream explosion that drives Mode 1a. The threshold is calibrated against the audit: BP and FA on no-LN architectures stay below $10$, while DFA on every audited architecture exceeds $90$.

\textbf{Diagnostic 2 (Reference validity).} Compute the deepest BP reference gradient norm $\|g_L\|$ at the end of training. Flag the run if $\|g_L\| < 10 \cdot \varepsilon$, where $\varepsilon$ is the cosine implementation's denominator floor for the training dtype (PyTorch's default is $\varepsilon = 10^{-8}$ for fp32, giving a threshold of $10^{-7}$ in our setup). At this threshold, the floor contributes more than $\sim 10\%$ of the cosine denominator, making the reported angle a mixture of the true reference direction and the implementation's clamp. In the audit, BP and FA on ResMLP $d=256, L=4$ stay above $10^{-6}$, while DFA collapses to $\sim 5 \times 10^{-10}$, well below the threshold.\looseness=-1

\textbf{Diagnostic 3 (Depth utility).} Compute the gap between the trained model's test accuracy and an architecture-matched frozen-blocks baseline (residual blocks frozen at random initialization, embedding/LN/head trained). Flag the run if the gap is below $2$ percentage points. This detects whether the trained depth contributes to the network's prediction, the question that Section~\ref{subsec:cosine-insufficient} showed cosine cannot answer even when measured validly. \emph{Precondition:} Diagnostic 3 is informative only when BP itself passes the frozen-blocks comparison on the same architecture and task, confirming that the task benefits from learned depth; when the frozen baseline already matches or exceeds BP, D3 cannot distinguish training failures from task-level properties (Appendix~\ref{app:ablation-extended}).

\textbf{Verdict logic.} A run fails the protocol if either Mode 1 (Diagnostics 1 and 2 both flag) or Depth Utility (Diagnostic 3 flags) is triggered. Diagnostic 1 alone does not constitute a Mode 1 failure---scale growth without reference collapse leaves cosine measurable, as observed for FA on ResMLP. Diagnostic 3 stands on its own: a run can pass both Mode 1 diagnostics and still fail at depth utility, which is precisely what Section~\ref{subsec:cosine-insufficient} establishes.\looseness=-1

\subsection{Decision-utility ablation}
\label{subsec:decision-utility}

To verify that each diagnostic group contributes to detecting failures the previous strategies miss, we ablate the protocol on two representative cases. Each case is a method-architecture pair where the standard reporting pair gives a misleading verdict, and we ask which reporting strategy first identifies the failure. Case 1 is DFA on ResMLP $d=256, L=4$ (the primary audit), where the failure is driven by Mode 1 measurement degeneracy. Case 2 is FA on ResMLP $d=512, L=2$ (the representative setting from Section~\ref{sec:motivating}), where Mode 1 is not triggered but the trained depth fails to outperform the frozen baseline. Together they exercise both the Mode 1 check and the depth utility check independently.\looseness=-1

\begin{table}[h]
    \centering
    \small
    \setlength{\tabcolsep}{6pt}
    \caption{Decision-utility ablation. Each cell shows the reporting strategy's verdict on the case; \textcolor{red!70!black}{\ding{55}} marks a missed failure. Both cases are genuine failures (the trained network fails to exceed a frozen-blocks baseline).}
    \label{tab:decision-utility}
    \begin{tabular*}{\linewidth}{@{\extracolsep{\fill}}lcc@{}}
        \toprule
        Reporting strategy & Case 1: DFA, $d=256, L=4$ & Case 2: FA, $d=512, L=2$ \\
        \midrule
        S0: acc only & \textcolor{red!70!black}{\ding{55}} pass (acc $0.306$) & \textcolor{red!70!black}{\ding{55}} pass (acc $0.345$) \\
        S1: acc + aggregate $\Gamma$ & \textcolor{red!70!black}{\ding{55}} pass ($\Gamma = +0.10$) & \textcolor{red!70!black}{\ding{55}} pass ($\Gamma = +0.48$) \\
        S2: + Mode 1 check (D1 \& D2) & \textcolor{green!50!black}{\checkmark} fail & \textcolor{red!70!black}{\ding{55}} pass (D1, D2 both ok) \\
        S3: + Depth utility (D3) & \textcolor{green!50!black}{\checkmark} fail & \textcolor{green!50!black}{\checkmark} fail \\
        \bottomrule
    \end{tabular*}
\end{table}

Table~\ref{tab:decision-utility} reports the verdicts. The standard pair (S0, S1) misses both failures. Adding the Mode 1 check (S2) catches Case 1 (DFA's reference collapses to $\sim 5\times 10^{-10}$) but not Case 2 (FA's reference remains valid). The depth utility check (S3) catches Case 2: FA's accuracy of $0.345$ fails to exceed the frozen baseline of $0.349$. The full protocol catches both.

Each of the three checks catches at least one case the others miss. Appendix~\ref{app:ablation-extended} extends this ablation to additional architectures and methods, with consistent results.

\subsection{Protocol verdict on CIFAR-100}
\label{subsec:cifar100}

To check whether the protocol's verdict pattern depends on the audited dataset, we replicate the primary-audit setup on CIFAR-100: ResMLP $d=256, L=4$, three seeds, 100 epochs, with the same optimizer, schedule, and training implementation (Section~\ref{sec:motivating}). We also compute an architecture-matched frozen-blocks baseline on CIFAR-100, which reaches $0.178 \pm 0.001$---close to the linear-probe ceiling on CIFAR-100 pixels ($0.177$).

\begin{table}[h]
    \centering
    \small
    \setlength{\tabcolsep}{6pt}
    \caption{Protocol verdict on CIFAR-100 (ResMLP $d=256, L=4$, three seeds, 100 epochs). The pattern from the primary CIFAR-10 audit is reproduced: BP passes; DFA fails Mode 1 and Depth Utility; FA passes Mode 1 (growth $15.6\times$, $\|g_L\| = 1.3 \times 10^{-6}$ above clamp) but fails Depth Utility ($-4.5$ pp vs frozen).}
    \label{tab:cifar100}
    \begin{tabular*}{\linewidth}{@{\extracolsep{\fill}}lcccccc@{}}
        \toprule
        Method & Test acc & D1 (Growth) & D2 ($\|g_L\|$) & D3 (vs Frozen) & Verdict \\
        \midrule
        BP  & $0.321 \pm 0.002$ & $1.0\times$ \textcolor{green!50!black}{\checkmark} & $9.5 \times 10^{-4}$ \textcolor{green!50!black}{\checkmark} & $+14.3$ pp \textcolor{green!50!black}{\checkmark} & \textcolor{green!50!black}{pass} \\
        FA  & $0.133 \pm 0.013$ & $15.6\times$ \textcolor{green!50!black}{\checkmark} & $1.3 \times 10^{-6}$ \textcolor{green!50!black}{\checkmark} & $-4.5$ pp \textcolor{red!70!black}{\ding{55}} & \textcolor{red!70!black}{fail (D3)} \\
        DFA & $0.088 \pm 0.001$ & $1004\times$ \textcolor{red!70!black}{\ding{55}} & $9.1 \times 10^{-9}$ \textcolor{red!70!black}{\ding{55}} & $-9.0$ pp \textcolor{red!70!black}{\ding{55}} & \textcolor{red!70!black}{fail (D1+D2+D3)} \\
        \bottomrule
    \end{tabular*}
\end{table}

As Table~\ref{tab:cifar100} shows, the CIFAR-100 verdict pattern matches CIFAR-10 exactly. BP passes all three diagnostics; DFA triggers Mode 1 (both D1 and D2) and depth utility, with the same residual-scale-explosion-to-reference-collapse mechanism developed in Section~\ref{sec:failure-modes}; FA's Mode 1 stays mild (growth below $50\times$, $\|g_L\|$ above the clamp) but its trained depth fails to outperform random blocks. FA on CIFAR-100 is a second instance of the depth-utility-only failure pattern (FA on $d=512, L=2$ ResMLP being the first in Section~\ref{subsec:decision-utility}), strengthening the case that Diagnostic 3 catches a class of failures distinct from those caught by Mode 1 alone. The protocol verdict is dataset-invariant on this architecture.

\subsection{Scope, recommendations, and discussion}
\label{subsec:scope}

\textbf{Scope.} The thresholds are calibrated on supervised image classification with small to medium residual architectures (ResMLP, ViT-Mini, StudentNet) on CIFAR-10 in fp32. On other architecture families, datasets, or precisions (fp16/bf16 raise the effective denominator floor), the threshold values may need recalibration (Appendix~\ref{app:threshold} reports their sensitivity on the audited runs); the measurements themselves remain well-defined wherever a forward pass, a BP reference gradient, and a frozen-blocks baseline can be computed. The Mode 1b mechanism is similarly scoped to terminal-LayerNorm architectures; on architectures without it, Diagnostic 2 functions as a defensive check rather than a primary failure detector. Extending the audit to larger models and datasets, to additional depth-utility baselines, to norm-control mechanisms gentler than the capacity-reducing penalty, and to RL, generative, or larger-scale NLP training is open work. Whether alignment measures beyond cosine---gradient signal-to-noise or Fisher-based proxies---add diagnostic value over per-layer cosine is likewise open.\looseness=-1

\textbf{Cross-architecture footprint.} Diagnostic 1 applies anywhere the local loss form $-\langle f_l(h_l), a_l \rangle$ is used without a norm penalty. Diagnostic 2 is informative only on architectures where the BP reference can mechanistically collapse; on no-LN architectures it adds little signal but costs nothing. Diagnostic 3 applies to any architecture where BP itself benefits from depth (i.e., BP passes the frozen-blocks comparison); on tasks solvable by a linear probe on random features, D3 is vacuously triggered for all methods including BP and should not be applied (Appendix~\ref{app:ablation-extended}).\looseness=-1

\textbf{Recommendations.} (i)~Report raw diagnostic values, not only pass/fail, so the thresholds can be recalibrated across toolchains and precisions. (ii)~Report cosine alignment per-layer rather than aggregate. (iii)~Treat the protocol as a minimum bar: passing rules out the failures we audited but does not validate the method as a BP alternative. (iv)~Avoid the measurement pitfalls catalogued in Appendix~\ref{app:pitfalls}.

%% file: sections/conclusion.tex
\section{Conclusion}
\label{sec:conclusion}

The standard reporting pair for feedback alignment---task accuracy and aggregate cosine alignment to the BP gradient---can fail to indicate that a method has trained the network, potentially leading researchers to build on or scale methods whose deep layers contribute nothing to the prediction.
%\cheng{Explain the negative consequence of "failing to indicate..."}
We identified two independent ways the cosine half can fail: the BP reference gradient can collapse below the cosine implementation's denominator floor in terminal-LayerNorm residual architectures (Mode 1), and the aggregate can mask layerwise heterogeneity that concentrates credit at a single end of the network (Mode 2). We further showed that even when cosine is measured validly, it does not predict whether trained depth contributes to the network's prediction; this is a separate failure of the accuracy half, requiring an architecture-matched frozen-blocks baseline to detect. The two failures are causally independent (Section~\ref{subsec:penalty}).\looseness=-1

%\cheng{it's unclear what "the recommended protocol" means exaclty. It's a protocol for what? Why do we need it? What's the use of it?  }
To enable practitioners to detect these silent failures before deploying or extending FA methods, the recommended protocol consists of three diagnostic checks---scale stability, reference validity, and depth utility---together with per-layer rather than aggregate cosine reporting. Across the audited architectures and methods, the standard reporting pair gives no signal of failure where our protocol identifies failure; passing all three diagnostics leaves the standard pair's interpretation intact but does not certify a method as universally effective.\looseness=-1

%\cheng{Overall, I feel that the paper has excellent content, but the high-level research story doesn't sound so compelling particularly because it's hard to know what exactly is the pain point that your work has addressed. Adding more discussion at the beginning of the paper to very clearly explain the pain point and argue why this is a serious pain point would help. It would also be necessary to then explain (later) how exactly your proposed methods or your findings have address the pain point. Think about how to explain this to someone who only builds agents on top of foundation model but doesn't train any LLM themselves. You need to assume that they don't really have direct experience with training any large NNs. If you can clearly explain your contribution to such a person, then you could see where you want add such an explanation in the paper, which may be scattered in different places of the paper, but you definitely want to have such sentences here or there to ensure successful communication of the value of the work to such a reviewer.  }

%% file: appendice/reproducibility.tex
\section{Reproducibility and experimental details}
\label{app:reproducibility}

Every number the paper reports is read from a saved result file. No model is retrained at reporting time. The reproduction notebook (\texttt{reproduce\_all.ipynb}) maps each table and figure to its source file. All runs use the seed set $\{42, 123, 456\}$, and every reported $\pm$ is a sample standard deviation over those seeds (ddof $=1$).

\subsection{Architectures}
\label{app:arch-specs}

\begin{table}[h]
    \centering
    \small
    \setlength{\tabcolsep}{6pt}
    \caption{Audited architectures. ResMLP is the primary backbone; the others test cross-architecture scope. StudentNet is trained on a synthetic teacher-student task described in the text.}
    \label{tab:arch-specs}
    \begin{tabular*}{\linewidth}{@{\extracolsep{\fill}}llcclc@{}}
        \toprule
        Architecture & Type & $d$ & $L$ & Normalization & Dataset \\
        \midrule
        ResMLP        & pre-LN residual MLP    & $256, 512$ & $2$--$12$ & block + terminal LN & CIFAR-10/100 \\
        ResMLP, no-LN & residual MLP           & $256$      & $4$       & block LN only       & CIFAR-10 \\
        ViT-Mini      & pre-LN transformer     & $128$      & $4$       & block + terminal LN & CIFAR-10 \\
        SmallCNN      & post-activation ResNet & $64$       & $4$       & BatchNorm           & CIFAR-10 \\
        StudentNet    & residual MLP           & $128$      & $4$       & none                & synthetic \\
        \bottomrule
    \end{tabular*}
\end{table}

ResMLP embeds a flattened CIFAR image with a linear map, applies $L$ pre-LayerNorm residual blocks $h_{l+1} = h_l + W_2\,\mathrm{GELU}(W_1\,\mathrm{LN}(h_l))$ with $W_1, W_2 \in \mathbb{R}^{d\times d}$ and $W_2$ initialized at $\mathcal{N}(0, 0.01^2)$, then a terminal LayerNorm and a linear head. The no-LN control removes the terminal LayerNorm; the no-residual ablation (Appendix~\ref{app:no-residual}) replaces each block by $h_{l+1}=W_2\,\mathrm{GELU}(W_1\,\mathrm{LN}(h_l))$ with $W_2$ initialized at standard deviation $0.5$. ViT-Mini embeds $4{\times}4$ patches into $64$ tokens plus a class token with learned positional embeddings, applies pre-LayerNorm transformer blocks ($4$ heads, MLP ratio $4$, GELU), and classifies the class token after a terminal LayerNorm. SmallCNN uses a $3{\times}3$ convolutional stem with BatchNorm, four post-activation residual blocks at width $64$, global average pooling, and a linear head. It has no LayerNorm. StudentNet is a residual MLP at $d{=}128$, $L{=}4$ with no embedding and no terminal LayerNorm; a fixed teacher (random $d{\times}d$ block weights spectrally normalized to $0.3$, $\tanh$ nonlinearity, seed $0$) labels standard-Gaussian $128$-dimensional inputs by argmax over $10$ classes, with $12{,}800$ training and $2{,}000$ test points.

\subsection{Training and credit rules}
\label{app:training-details}

The shared recipe is AdamW at learning rate $10^{-3}$, weight decay $0.01$, a cosine schedule, batch size $128$, and $100$ epochs; penalty and dissociation runs use $30$ epochs. The synthetic StudentNet task uses batch size $256$. Each non-BP method updates block $f_l$ by the local loss $-\langle f_l(h_l), a_l\rangle$, and the output head is trained by the exact cross-entropy gradient on a detached $h_L$. FA fixes $B_l \in \mathbb{R}^{d\times d}$ at $\mathcal{N}(0,1)/\sqrt{d}$; DFA fixes $B_l \in \mathbb{R}^{C\times d}$. State Bridge and Credit Bridge each use a three-layer MLP (hidden width $256$, GELU) over $[\mathrm{LN}(h_l),\ \text{time-embed}(l/L),\ s]$ with a sinusoidal time embedding of dimension $32$ and an EMA target at momentum $0.995$; State Bridge predicts $h_L$ and Credit Bridge outputs a scalar value. The per-block scale penalty, when used, adds $\lambda\,\|f_l(h_l)\|^2$ to each block's local loss at $\lambda \in \{10^{-4}, 10^{-2}\}$.

\subsection{Measurement conventions}
\label{app:measurement}

Diagnostics are measured at the final checkpoint, or along a stored per-epoch trajectory for the temporal figures. Per-layer cosines and gradient norms use an evaluation batch of $2048$ CIFAR-10 test inputs with the model in eval mode, and the feedback matrices $B_l$ reconstructed from the training RNG. The layer index convention is $l{=}0$ for the first residual block.

\subsection{Source of reported values}
\label{app:source-map}

The reported tables and figures derive from saved files under \texttt{results/}: the audit from \texttt{protocol\_audit/audit\_table\_s42\_s123\_s456.json}, the penalty rescue from the \texttt{round38} and \texttt{round41} runs, the depth sweep from \texttt{fa\_depth\_scan\_d512\_L*} and \texttt{cifar\_depth\_scan\_s42}, the no-residual ablation from \texttt{h2\_no\_residual\_full\_s*}, the random-target ablation from the \texttt{*\_random\_targets\_*} runs, the layer-0 table from \texttt{vanilla\_dfa\_early\_ckpts}, the threshold sweep from \texttt{threshold\_sensitivity\_output.txt}, and the depth-utility ladder from \texttt{depth\_ladder/ladder\_*.json}. The reproduction notebook (\texttt{reproduce\_all.ipynb}) lists the complete map.

%% file: appendice/per-arch.tex
\section{Per-architecture detailed audit}
\label{app:per-arch}

Diagnostic 2 (reference collapse) fires only on the terminal-LayerNorm architectures (Table~\ref{tab:per-arch}). On ResMLP and ViT-Mini, FA and DFA drive $\|g_L\|$ to the floor; on the no-LN ResMLP and StudentNet it holds at $10^{-4}$ or above even where growth $\rho$ exceeds $300$, so the protocol does not flag those runs for Mode 1. Growth and $\|g_L\|$ follow the conventions of Appendix~\ref{app:reproducibility}. Cross-architecture rows are seed 42.

\begin{table}[h]
    \centering
    \small
    \setlength{\tabcolsep}{4pt}
    \caption{Per-architecture protocol audit. D1 flags growth $\rho>50$, D2 flags $\|g_L\|<10^{-7}$, D3 flags a frozen-baseline gap below $2$ points. A run fails only if Mode 1 (D1 and D2) or D3 triggers; D1 alone is not a failure. D3 is n/a on StudentNet (frozen $>$ BP) and on the no-LN control (not assessed). \textcolor{green!50!black}{\checkmark} passes, \textcolor{red!70!black}{\ding{55}} flags.}
    \label{tab:per-arch}
    \begin{tabular*}{\linewidth}{@{\extracolsep{\fill}}lccccc@{}}
        \toprule
        Method & Test acc & D1 (growth $\rho$) & D2 ($\|g_L\|$) & D3 (vs frozen) & Verdict \\
        \midrule
        \multicolumn{6}{@{}l}{\emph{ResMLP $d{=}256, L{=}4$, CIFAR-10 (three seeds); frozen $0.349$}} \\
        BP  & $0.615$ & $1.0\times$ \textcolor{green!50!black}{\checkmark} & $3.6\times10^{-4}$ \textcolor{green!50!black}{\checkmark} & $+26.6$ \textcolor{green!50!black}{\checkmark} & \textcolor{green!50!black}{pass} \\
        FA  & $0.401$ & $13\times$ \textcolor{green!50!black}{\checkmark} & $1.2\times10^{-6}$ \textcolor{green!50!black}{\checkmark} & $+5.2$ \textcolor{green!50!black}{\checkmark} & \textcolor{green!50!black}{pass} \\
        DFA & $0.306$ & $1856\times$ \textcolor{red!70!black}{\ding{55}} & $3.0\times10^{-9}$ \textcolor{red!70!black}{\ding{55}} & $-4.3$ \textcolor{red!70!black}{\ding{55}} & \textcolor{red!70!black}{fail} \\
        \addlinespace[2pt]
        \multicolumn{6}{@{}l}{\emph{ViT-Mini $d{=}128, L{=}4$ (seed 42); frozen $0.570$}} \\
        BP  & $0.799$ & $4.1\times$ \textcolor{green!50!black}{\checkmark} & $7.1\times10^{-6}$ \textcolor{green!50!black}{\checkmark} & $+22.9$ \textcolor{green!50!black}{\checkmark} & \textcolor{green!50!black}{pass} \\
        FA  & $0.163$ & $2.5\times10^{7}$ \textcolor{red!70!black}{\ding{55}} & $5.1\times10^{-11}$ \textcolor{red!70!black}{\ding{55}} & $-40.7$ \textcolor{red!70!black}{\ding{55}} & \textcolor{red!70!black}{fail} \\
        DFA & $0.256$ & $7.6\times10^{5}$ \textcolor{red!70!black}{\ding{55}} & $2.6\times10^{-10}$ \textcolor{red!70!black}{\ding{55}} & $-31.4$ \textcolor{red!70!black}{\ding{55}} & \textcolor{red!70!black}{fail} \\
        \addlinespace[2pt]
        \multicolumn{6}{@{}l}{\emph{ResMLP, no terminal LN, $d{=}256, L{=}4$ (seed 42)}} \\
        BP  & $0.616$ & $0.8\times$ \textcolor{green!50!black}{\checkmark} & $1.2\times10^{-4}$ \textcolor{green!50!black}{\checkmark} & n/a & \textcolor{green!50!black}{pass} \\
        FA  & $0.295$ & $323\times$ \textcolor{red!70!black}{\ding{55}} & $6.4\times10^{-4}$ \textcolor{green!50!black}{\checkmark} & n/a & \textcolor{green!50!black}{pass} \\
        DFA & $0.332$ & $1380\times$ \textcolor{red!70!black}{\ding{55}} & $7.4\times10^{-4}$ \textcolor{green!50!black}{\checkmark} & n/a & \textcolor{green!50!black}{pass} \\
        \addlinespace[2pt]
        \multicolumn{6}{@{}l}{\emph{StudentNet $d{=}128, L{=}4$ (seed 42); frozen $0.908 >$ BP}} \\
        BP  & $0.796$ & $2.0\times$ \textcolor{green!50!black}{\checkmark} & $1.2\times10^{-6}$ \textcolor{green!50!black}{\checkmark} & n/a & \textcolor{green!50!black}{pass} \\
        FA  & $0.414$ & $64\times$ \textcolor{red!70!black}{\ding{55}} & $3.1\times10^{-4}$ \textcolor{green!50!black}{\checkmark} & n/a & \textcolor{green!50!black}{pass} \\
        DFA & $0.693$ & $6.9\times$ \textcolor{green!50!black}{\checkmark} & $1.2\times10^{-4}$ \textcolor{green!50!black}{\checkmark} & n/a & \textcolor{green!50!black}{pass} \\
        \bottomrule
    \end{tabular*}
\end{table}

Diagnostic 3 applies only where BP beats the frozen baseline. ResMLP and ViT-Mini qualify, and the trained FA and DFA blocks fall below the baseline there. StudentNet does not qualify: its frozen baseline ($0.908$) exceeds even BP ($0.796$). FA passes all three diagnostics on the primary ResMLP audit. On ViT-Mini, which also carries terminal LayerNorm, FA collapses alongside DFA.

%% file: appendice/depth-sweep.tex
\section{Depth-sweep layerwise profiles}
\label{app:depth-sweep}

Across depths $L\in\{2,4,6,8,12\}$, DFA's mean deep-block BP cosine stays within $[-0.005,+0.000]$, FA's runs from $+0.96$ at $L{=}2$ to $+0.09$ at $L{=}12$, and BP's holds at $+0.94$ or above (Table~\ref{tab:depth-sweep}). We ran the $d{=}512$ pre-LayerNorm ResMLP of Section~\ref{sec:motivating} at these five depths on CIFAR-10 (seed 42, otherwise matched); each row reports the layer-0 BP cosine, the mean BP cosine over the $L{-}1$ deeper blocks, and the matching perturbation correlation $\rho$. The reference stays measurable at every depth.

\begin{table}[h]
    \centering
    \small
    \setlength{\tabcolsep}{6pt}
    \caption{Depth sweep on $d{=}512$ ResMLP, seed 42, 100 epochs, CIFAR-10. \emph{layer-0 cos}: embedding-block BP cosine. \emph{deep cos}: mean BP cosine over the remaining $L{-}1$ blocks. \emph{deep $\rho$}: matching mean perturbation correlation.}
    \label{tab:depth-sweep}
    \begin{tabular*}{\linewidth}{@{\extracolsep{\fill}}rlcccc@{}}
        \toprule
        $L$ & Method & Test acc & layer-0 cos & deep cos & deep $\rho$ \\
        \midrule
        $2$ & BP            & $0.599$ & $+1.000$ & $+1.000$ & $+0.983$ \\
        $2$ & FA            & $0.349$ & $+0.034$ & $+0.956$ & $+0.051$ \\
        $2$ & DFA           & $0.312$ & $+0.396$ & $-0.005$ & $+0.000$ \\
        $2$ & Credit Bridge & $0.310$ & $+0.330$ & $+0.020$ & $+0.000$ \\
        \addlinespace[2pt]
        $4$ & BP            & $0.603$ & $+1.000$ & $+1.000$ & $+0.988$ \\
        $4$ & FA            & $0.424$ & $+0.007$ & $+0.289$ & $-0.009$ \\
        $4$ & DFA           & $0.314$ & $+0.400$ & $-0.000$ & $+0.000$ \\
        $4$ & Credit Bridge & $0.298$ & $+0.402$ & $+0.030$ & $+0.000$ \\
        \addlinespace[2pt]
        $6$ & BP            & $0.602$ & $+0.993$ & $+0.993$ & $+0.991$ \\
        $6$ & FA            & $0.401$ & $+0.019$ & $+0.162$ & $-0.005$ \\
        $6$ & DFA           & $0.310$ & $+0.387$ & $-0.000$ & $+0.000$ \\
        $6$ & Credit Bridge & $0.299$ & $+0.304$ & $+0.054$ & $+0.000$ \\
        \addlinespace[2pt]
        $8$ & BP            & $0.589$ & $+0.965$ & $+0.965$ & $+0.992$ \\
        $8$ & FA            & $0.409$ & $+0.026$ & $+0.115$ & $-0.003$ \\
        $8$ & DFA           & $0.306$ & $+0.377$ & $-0.000$ & $+0.000$ \\
        $8$ & Credit Bridge & $0.288$ & $+0.205$ & $+0.022$ & $+0.000$ \\
        \addlinespace[2pt]
        $12$ & BP            & $0.594$ & $+0.942$ & $+0.940$ & $+0.990$ \\
        $12$ & FA            & $0.404$ & $+0.046$ & $+0.091$ & $+0.013$ \\
        $12$ & DFA           & $0.309$ & $+0.388$ & $-0.000$ & $+0.000$ \\
        $12$ & Credit Bridge & $0.239$ & $+0.208$ & $+0.016$ & $+0.000$ \\
        \bottomrule
    \end{tabular*}
\end{table}

DFA's layer-0 cosine holds in $[+0.38,+0.40]$ at all five depths, and its deep $\rho$ stays at numerical zero. FA shows the inverse profile, with layer-0 cosine near zero ($\le +0.05$) and a deep cosine carried by its deepest block ($+0.995$ at the final block of $L{=}4$). BP keeps a deep cosine of $+0.94$ at $L{=}12$. Credit Bridge has a small positive deep cosine ($\le +0.054$) that does not grow as depth falls.

The $L{=}4$ row, replicated over seeds 42, 123, and 456, gives DFA layer-0 $+0.412\pm0.013$, DFA deep $-0.0004\pm0.0009$, and CB deep $+0.039\pm0.012$. Shrinking the network does not move DFA credit into the deep blocks. FA's accuracy exceeds DFA's at every depth, $0.404$ against $0.309$ at $L{=}12$. The layerwise pattern is depth-invariant.

%% file: appendice/ablations.tex
\section{No-residual and random-target ablations}
\label{app:ablations}

Residual-scale growth under DFA does not require the additive skip or any task signal. Both ablations use the 4-block $d{=}256$ pre-LayerNorm ResMLP and the recipe of Section~\ref{sec:motivating}, seed 42. Diagnostic~1 and Diagnostic~2 are defined in Section~\ref{subsec:diagnostics}.

\subsection{No-residual ablation}
\label{app:no-residual}

Without the additive skip, DFA's residual stream still grows. We replace each block $h_{l+1}=h_l+f_l(h_l)$ with $h_{l+1}=f_l(h_l)$ and raise the inner-layer initialization standard deviation from $0.01$ to $0.5$ so the stack trains from step zero; terminal LayerNorm is unchanged. We test DFA, where the residual growth is largest, with BP as a control. In three epochs DFA's $\|h_L\|$ rises from $4.69$ to $2.2\times10^4$ and $\|g_L\|$ falls from $9.8\times10^{-4}$ to $1.6\times10^{-7}$ (Table~\ref{tab:no-residual}). Diagnostic~1 fires and Diagnostic~2 reaches the floor. The skip is not necessary for the growth.

\begin{table}[h]
    \centering
    \small
    \setlength{\tabcolsep}{6pt}
    \caption{No-residual ResMLP-$d256$, seed 42, three epochs. Each block is $h_{l+1}=f_l(h_l)$ with inner-init std $0.5$; terminal LayerNorm unchanged. $\Gamma_{\mathrm{DFA}}$ is the aggregate DFA cosine (n/a for BP and at initialization).}
    \label{tab:no-residual}
    \begin{tabular*}{\linewidth}{@{\extracolsep{\fill}}lcccccc@{}}
        \toprule
        Method & $w_2$ std & ep & $\|h_L\|$ & $\|g_L\|$ & Test acc & $\Gamma_{\mathrm{DFA}}$ \\
        \midrule
        BP  & $0.5$ & $0$ & $4.69$     & $9.8\times 10^{-4}$ & $0.080$ & n/a \\
        BP  & $0.5$ & $1$ & $155$      & $4.3\times 10^{-5}$ & $0.144$ & n/a \\
        BP  & $0.5$ & $2$ & $174$      & $4.0\times 10^{-5}$ & $0.164$ & n/a \\
        BP  & $0.5$ & $3$ & $163$      & $4.2\times 10^{-5}$ & $0.163$ & n/a \\
        \addlinespace[2pt]
        DFA & $0.5$ & $0$ & $4.69$     & $9.8\times 10^{-4}$ & $0.080$ & n/a \\
        DFA & $0.5$ & $1$ & $5{,}295$  & $8.6\times 10^{-7}$ & $0.156$ & $0.047$ \\
        DFA & $0.5$ & $2$ & $16{,}930$ & $2.2\times 10^{-7}$ & $0.151$ & $0.040$ \\
        DFA & $0.5$ & $3$ & $22{,}050$ & $1.6\times 10^{-7}$ & $0.148$ & $0.039$ \\
        \bottomrule
    \end{tabular*}
\end{table}

The converged run shows the same trend. Over 100 epochs and seeds 42, 123, and 456 the no-residual run reaches mean $\|h_L\|\approx 8.2\times10^7$ and $\|g_L\|\approx 1.9\times10^{-10}$, within an order of magnitude of vanilla residual DFA on this backbone ($5\times10^8$ and $4\times10^{-10}$). Inner-init std $\{0.1,0.2,0.5\}$ give the same trend. BP learns only at std $0.5$; this bounds the algorithm comparison, and the growth observation stands either way.

\subsection{Random-target ablation}
\label{app:random-target}

With random labels the network learns nothing, and the residual stream still inflates. We retrain DFA on CIFAR-10 inputs with i.i.d.\ class labels redrawn each minibatch from a uniform distribution over $\{0,\dots,9\}$; $B_l$ and all hyperparameters match the Section~\ref{sec:motivating} run. Test accuracy stays at chance ($0.07$ at epoch 3). In three epochs $\|h_L\|$ grows from $8.89$ to $1.45\times10^4$ and $\|g_L\|$ falls from $9.8\times10^{-4}$ to $5.6\times10^{-7}$ (Table~\ref{tab:random-target}); the 100-epoch run reaches $1.67\times10^8$ and $8.0\times10^{-12}$. Task signal is not required for the growth.

\begin{table}[h]
    \centering
    \small
    \setlength{\tabcolsep}{6pt}
    \caption{Random-target DFA on the standard residual ResMLP-$d256$, seed 42, three epochs. Class labels are i.i.d.\ uniform over $\{0,\dots,9\}$, redrawn each minibatch. $\Gamma_{\mathrm{DFA}}$ is the aggregate DFA cosine (n/a at initialization).}
    \label{tab:random-target}
    \begin{tabular*}{\linewidth}{@{\extracolsep{\fill}}ccccc@{}}
        \toprule
        ep & $\|h_L\|$ & $\|g_L\|$ & Test acc & $\Gamma_{\mathrm{DFA}}$ \\
        \midrule
        $0$ & $8.89$     & $9.83\times 10^{-4}$ & $0.115$ & n/a \\
        $1$ & $1{,}616$  & $5.12\times 10^{-6}$ & $0.078$ & $-0.020$ \\
        $2$ & $9{,}768$  & $8.50\times 10^{-7}$ & $0.081$ & $-0.024$ \\
        $3$ & $14{,}510$ & $5.62\times 10^{-7}$ & $0.071$ & $-0.025$ \\
        \bottomrule
    \end{tabular*}
\end{table}

State Bridge and Credit Bridge show the same data-agnostic growth (Table~\ref{tab:random-target-sbcb}). At epoch 3 of random-target training their $\|h_L\|$ reaches $6.2\times10^3$ and $2.0\times10^4$ from a start near $9$, both at chance accuracy. Their $\|g_L\|$ stays above the Diagnostic~2 floor at this horizon, so only the growth half of Mode~1 has appeared. Both use bridge constructions with target normalization and stop-gradients. DFA, State Bridge, and Credit Bridge all show the growth. FA under random targets converges to $\|h_L\|\approx1.3\times10^{5}$ at chance accuracy, with $\|g_L\|\approx2.4\times10^{-7}$ above the Diagnostic~2 floor.

\begin{table}[h]
    \centering
    \small
    \setlength{\tabcolsep}{6pt}
    \caption{Random-target ablation at epoch 3, standard residual ResMLP-$d256$, seed 42, i.i.d.\ uniform class labels. All three audited fixed-feedback methods are shown.}
    \label{tab:random-target-sbcb}
    \begin{tabular*}{\linewidth}{@{\extracolsep{\fill}}lccc@{}}
        \toprule
        Method & $\|h_L\|$ at ep 3 & $\|g_L\|$ at ep 3 & Test acc \\
        \midrule
        DFA           & $14{,}510$ & $5.6\times 10^{-7}$ & $0.071$ \\
        State Bridge  & $6{,}225$  & $1.0\times 10^{-5}$ & $0.104$ \\
        Credit Bridge & $19{,}974$ & $3.2\times 10^{-6}$ & $0.092$ \\
        \bottomrule
    \end{tabular*}
\end{table}

%% file: appendice/layer0-dominance.tex
\section{Layer-0 dominance in vanilla DFA}
\label{app:layer0}

On vanilla DFA, layer 0 carries the aggregate cosine and the deep blocks carry almost none. We measured per-layer cosines between DFA's credit $a_l = B_l^\top e_T$ and the BP gradient on saved early-epoch checkpoints, at an eval batch of $n{=}2048$ CIFAR-10 test inputs with the training-time $B_l$. Index $l{=}0$ is the first residual block; $l{=}1,\dots,4$ are the deeper blocks. Table~\ref{tab:layer0} gives the values.

\begin{table}[h]
    \centering
    \small
    \setlength{\tabcolsep}{6pt}
    \caption{Per-layer cosines on vanilla DFA early-epoch checkpoints, three seeds, epochs 1 and 2, 4-block $d{=}256$ pre-LayerNorm ResMLP. $l{=}0$ is the first residual block; $l{=}1,\dots,4$ are the deeper blocks. $\|g_2\|$ is the layer-2 BP gradient norm.}
    \label{tab:layer0}
    \begin{tabular*}{\linewidth}{@{\extracolsep{\fill}}rrrrrrrr@{}}
        \toprule
        seed & ep & $l{=}0$ & $l{=}1$ & $l{=}2$ & $l{=}3$ & $l{=}4$ & $\|g_2\|$ \\
        \midrule
        $42$  & $1$ & $+0.421$ & $+0.005$ & $-0.028$ & $-0.039$ & $-0.038$ & $6.8\times10^{-7}$ \\
        $42$  & $2$ & $+0.437$ & $-0.002$ & $-0.040$ & $-0.055$ & $-0.054$ & $1.6\times10^{-7}$ \\
        \addlinespace[2pt]
        $123$ & $1$ & $+0.436$ & $+0.008$ & $-0.033$ & $+0.016$ & $+0.017$ & $6.6\times10^{-7}$ \\
        $123$ & $2$ & $+0.460$ & $+0.005$ & $-0.037$ & $+0.003$ & $+0.003$ & $1.4\times10^{-7}$ \\
        \addlinespace[2pt]
        $456$ & $1$ & $+0.418$ & $+0.011$ & $-0.026$ & $+0.007$ & $+0.006$ & $3.8\times10^{-7}$ \\
        $456$ & $2$ & $+0.409$ & $+0.003$ & $-0.039$ & $+0.001$ & $+0.000$ & $8.5\times10^{-8}$ \\
        \bottomrule
    \end{tabular*}
\end{table}

Layer 0 stays at $+0.42 \pm 0.02$ across all six measurements, while every deep layer lies in $[-0.06, +0.02]$. The deep-layer mean is $-0.008 \pm 0.016$ at epoch 1 and $-0.018 \pm 0.017$ at epoch 2. At epoch 1 every $\|g_2\|$ exceeds $3.8\times10^{-7}$, above the Diagnostic~2 floor of $10^{-7}$. The aggregate cosine here comes from layer 0. Mode 2 is present while the reference is still valid.

%% file: appendice/A.tex
\section{FA + penalty on ResMLP: no dose response}
\label{app:fa-penalty}

For completeness, we apply the same penalty intervention to FA on ResMLP $d=256, L=4$, three seeds, 30 epochs:

\begin{center}
\small
\setlength{\tabcolsep}{6pt}
\begin{tabular*}{\linewidth}{@{\extracolsep{\fill}}lcccc@{}}
\toprule
$\lambda$ & Test acc & Deep cos & $\|h_L\|$ & $\|g_L\|$ \\
\midrule
$0$       & $0.372 \pm 0.007$ & $+0.325 \pm 0.015$ & $\sim 9 \times 10^4$ & $\sim 2 \times 10^{-6}$ \\
$10^{-4}$ & $0.377 \pm 0.006$ & $+0.298 \pm 0.031$ & $\sim 9 \times 10^3$ & $\sim 1 \times 10^{-5}$ \\
$10^{-2}$ & $0.369 \pm 0.003$ & $+0.423 \pm 0.006$ & $\sim 1 \times 10^4$ & $\sim 2 \times 10^{-5}$ \\
\bottomrule
\end{tabular*}
\end{center}

FA on this architecture has mild Mode 1 even at $\lambda = 0$: the residual norm $\|h_L\| \sim 9 \times 10^4$ is well below DFA's $\sim 4 \times 10^8$, and the BP reference gradient $\|g_L\| \sim 2 \times 10^{-6}$ is already two orders of magnitude above the cosine clamp. The penalty has no Mode 1 to alleviate; accordingly, accuracy and deep cosine remain in narrow windows ($0.37$--$0.38$ and $+0.30$--$+0.42$ respectively) across all three $\lambda$ values. This is consistent with the cross-architecture finding (Section~\ref{subsec:mode1-scope}) that FA's Mode 1 severity depends on the architecture, and the intervention is most informative where Mode 1 is fully developed.

%% file: appendice/sbcb-rescue.tex
\section{State Bridge and Credit Bridge under penalty rescue}
\label{app:sbcb-rescue}

Under the per-block penalty $\lambda{=}10^{-2}$ that rescues DFA in Section~\ref{subsec:penalty}, State Bridge reaches $0.453 \pm 0.003$ test accuracy over three seeds, $9.3$ points above DFA+penalty and $10.4$ above the frozen-blocks baseline of $0.349$ (Table~\ref{tab:sbcb-rescue}). The penalty is added to each method's per-block local loss only; the bridge predictor, the value network, and the embedding and head stay unpenalized. State Bridge and Credit Bridge run on the 4-block $d{=}256$ pre-LayerNorm ResMLP for 30 epochs and seeds 42, 123, and 456. All other settings match Section~\ref{subsec:penalty}.

\begin{table}[h]
    \centering
    \small
    \setlength{\tabcolsep}{6pt}
    \caption{State Bridge and Credit Bridge under per-block penalty $\lambda{=}10^{-2}$, 4-block $d{=}256$ pre-LayerNorm ResMLP, 30 epochs. Per-seed rows and the mean ($\pm$ std over seeds 42, 123, 456, ddof $=1$). Vanilla rows ($\lambda{=}0$, seed 42) and the DFA+penalty mean are references. deep cos and deep $\rho$ are n/a where the reference gradient has collapsed.}
    \label{tab:sbcb-rescue}
    \begin{tabular*}{\linewidth}{@{\extracolsep{\fill}}lccccc@{}}
        \toprule
        Run & Test acc & $\|h_L\|$ & $\|g_L\|$ & deep cos & deep $\rho$ \\
        \midrule
        SB+pen, $s42$  & $0.456$ & $302$ & $1.75\times10^{-4}$ & $+0.312$ & $+0.392$ \\
        SB+pen, $s123$ & $0.451$ & $311$ & $1.74\times10^{-4}$ & $+0.327$ & $+0.424$ \\
        SB+pen, $s456$ & $0.451$ & $292$ & $1.92\times10^{-4}$ & $+0.326$ & $+0.391$ \\
        SB+pen, mean   & $0.453 \pm 0.003$ & $302$ & $1.80\times10^{-4}$ & $+0.322 \pm 0.008$ & $+0.402 \pm 0.019$ \\
        \addlinespace[2pt]
        CB+pen, $s42$  & $0.360$ & $5{,}431$ & $1.88\times10^{-5}$ & $+0.684$ & $+0.498$ \\
        CB+pen, $s123$ & $0.364$ & $5{,}834$ & $1.81\times10^{-5}$ & $+0.667$ & $+0.452$ \\
        CB+pen, $s456$ & $0.356$ & $5{,}775$ & $2.01\times10^{-5}$ & $+0.685$ & $+0.442$ \\
        CB+pen, mean   & $0.360 \pm 0.004$ & $5{,}680$ & $1.90\times10^{-5}$ & $+0.679 \pm 0.010$ & $+0.464 \pm 0.030$ \\
        \addlinespace[2pt]
        SB, $\lambda{=}0$  & $0.213$ & $9.85\times10^{6}$ & $1\times10^{-8}$ & n/a & n/a \\
        CB, $\lambda{=}0$  & $0.211$ & $6.7\times10^{7}$  & ${\sim}0$           & n/a & n/a \\
        DFA+pen, mean      & $0.360 \pm 0.002$ & $1.3\times10^{4}$ & $1.6\times10^{-6}$ & $+0.151 \pm 0.025$ & $+0.080 \pm 0.012$ \\
        \bottomrule
    \end{tabular*}
\end{table}

Credit Bridge has the higher deep cosine, $+0.679 \pm 0.010$ against State Bridge's $+0.322$, yet it only matches DFA's accuracy. Table~\ref{tab:sbcb-rescue} adds the per-seed values and the residual-scale and reference-norm columns that Section~\ref{subsec:cosine-insufficient} omits. Vanilla State Bridge and Credit Bridge ($\lambda{=}0$) reach $\|h_L\|$ of $9.85\times10^{6}$ and $6.7\times10^{7}$ at accuracies $0.213$ and $0.211$. The penalty is what contains the residual stream here.

%% file: appendice/B.tex
\section{Extended decision-utility ablation}
\label{app:ablation-extended}

We extend the decision-utility ablation to additional architecture-method pairs to verify that the Section~\ref{subsec:decision-utility} pattern (each diagnostic group catching at least one case the others miss) is not specific to the two cases shown in the main text.

\begin{table}[h]
    \centering
    \small
    \setlength{\tabcolsep}{4pt}
    \caption{Extended decision-utility ablation. Each cell shows the reporting strategy's verdict; \textcolor{red!70!black}{\ding{55}} marks a missed failure (silently passing a method whose trained blocks underperform a frozen-blocks baseline). Healthy reference cases (BP) appear with \textcolor{green!50!black}{\checkmark} throughout. Cases include the two main-text cases (Section~\ref{subsec:decision-utility}) plus three additional method-architecture pairs.}
    \label{tab:ablation-extended}
    \resizebox{\linewidth}{!}{%
    \begin{tabular}{lccccc}
        \toprule
        Reporting strategy & DFA, RM-d256-L4 & FA, RM-d512-L2 & DFA, ViT-Mini & FA, ViT-Mini & BP, RM-d256-L4 \\
        \midrule
        S0: acc only & \textcolor{red!70!black}{\ding{55}} & \textcolor{red!70!black}{\ding{55}} & \textcolor{red!70!black}{\ding{55}} & \textcolor{red!70!black}{\ding{55}} & \textcolor{green!50!black}{\checkmark} \\
        S1: acc + $\Gamma$ & \textcolor{red!70!black}{\ding{55}} & \textcolor{red!70!black}{\ding{55}} & \textcolor{red!70!black}{\ding{55}} & \textcolor{red!70!black}{\ding{55}} & \textcolor{green!50!black}{\checkmark} \\
        S2: + Mode 1 (D1 \& D2) & \textcolor{green!50!black}{\checkmark} & \textcolor{red!70!black}{\ding{55}} & \textcolor{green!50!black}{\checkmark} & \textcolor{green!50!black}{\checkmark} & \textcolor{green!50!black}{\checkmark} \\
        S3: + Depth utility (D3) & \textcolor{green!50!black}{\checkmark} & \textcolor{green!50!black}{\checkmark} & \textcolor{green!50!black}{\checkmark} & \textcolor{green!50!black}{\checkmark} & \textcolor{green!50!black}{\checkmark} \\
        \bottomrule
    \end{tabular}%
    }
\end{table}

As Table~\ref{tab:ablation-extended} shows, the pattern from the main text holds: the standard pair (S0, S1) silently passes all four genuine failure cases; the Mode 1 check (S2) catches the three cases where measurement is degenerate but does not catch the case where measurement is valid (FA on $d=512, L=2$); the depth utility check (S3) closes the remaining gap. The healthy case (BP on the primary audit) passes all reporting strategies, confirming that the protocol does not flag working methods. ViT-Mini provides the strongest depth-utility evidence in the extended set: the frozen-blocks baseline reaches $0.570 \pm 0.003$ on this architecture, while DFA and FA reach $0.256$ and $0.163$ respectively---trained transformer blocks underperform random blocks by more than $30$ percentage points.

StudentNet, the no-LN architecture used in Section~\ref{subsec:mode1-scope} as a Mode 1b control, is not included in this ablation because it does not satisfy the precondition for Diagnostic~3. Diagnostic~3 is informative only when the task actually benefits from learned depth---operationally, when BP itself passes the frozen-blocks comparison. On StudentNet, the task is solvable by a linear probe on random features: the frozen-blocks baseline reaches $0.908 \pm 0.009$, exceeding even fully trained BP ($0.796$). In this regime, every method---including BP---would be flagged by D3, so the diagnostic cannot distinguish training failures from task-level properties. We use StudentNet only as a mechanism control for Mode 1b's architectural scope and not as a depth-utility test case.

%% file: appendice/depth-ladder.tex
\section{Depth-utility ladder}
\label{app:depth-ladder}

\begin{figure}[h]
    \centering
    \includegraphics[width=\linewidth]{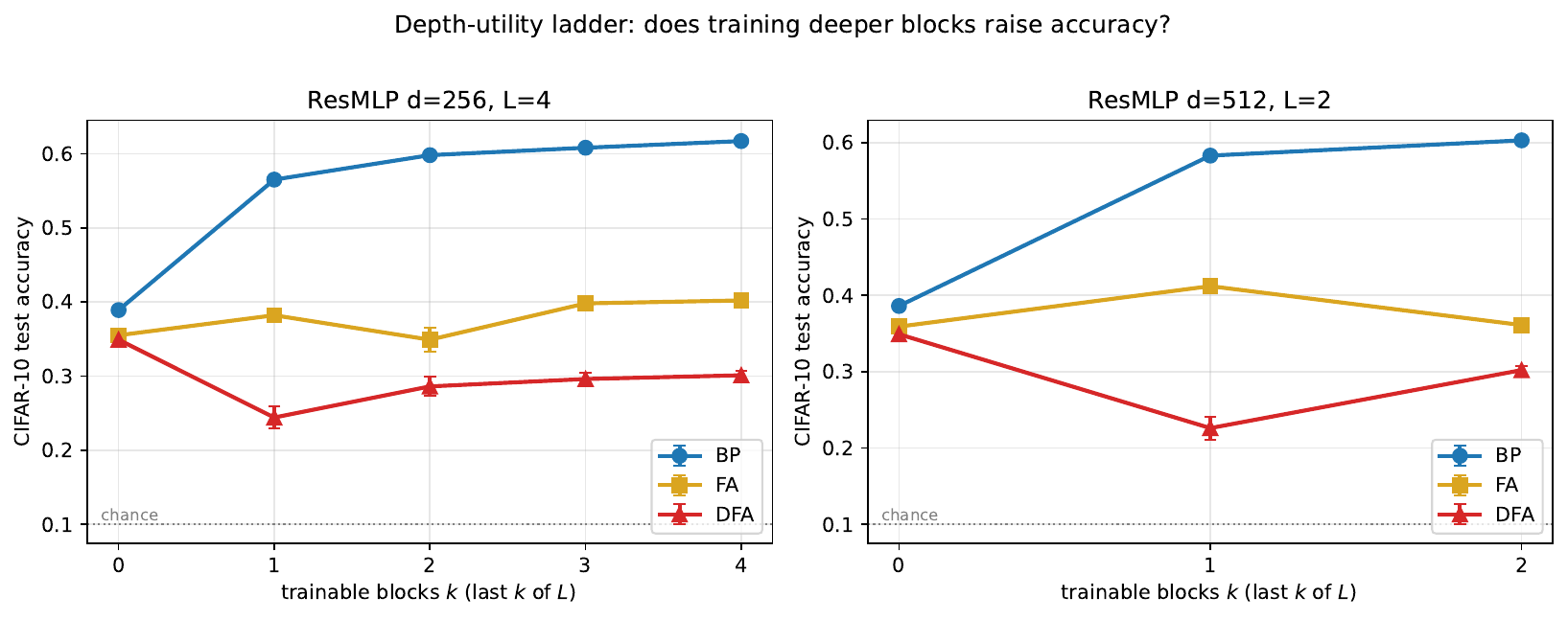}
    \caption{Depth-utility ladder: CIFAR-10 test accuracy versus the number of trainable blocks $k$ (last $k$ of $L$), three seeds. $k{=}0$ is the frozen-blocks baseline, $k{=}L$ the full audit. BP climbs; DFA stays at or below the frozen rung; FA is intermediate.}
    \label{fig:depth-ladder}
\end{figure}

BP's accuracy climbs monotonically as more residual blocks are made trainable, while DFA's peaks at the frozen baseline and falls once any deep block trains (Figure~\ref{fig:depth-ladder}). We train the last $k$ blocks and freeze the first $L{-}k$ at random initialization. The embedding, LayerNorm, and head are always trained. Credit still propagates through the frozen blocks; only their weights stay fixed, so $k{=}0$ recovers the frozen-blocks baseline and $k{=}L$ the full audit. We sweep $k$ for BP, FA, and DFA on ResMLP $d{=}256, L{=}4$ and $d{=}512, L{=}2$ over three seeds, under the recipe of Appendix~\ref{app:reproducibility}. Training the last $k$ blocks is the best case for each method: the deepest block receives the most direct credit.

\begin{table}[h]
    \centering
    \small
    \setlength{\tabcolsep}{6pt}
    \caption{Depth-utility ladder: CIFAR-10 test accuracy (mean $\pm$ ddof-1 std, three seeds) versus the number of trainable blocks $k$. $k{=}0$ is the frozen-blocks baseline, $k{=}L$ the full audit.}
    \label{tab:depth-ladder}
    \begin{tabular*}{\linewidth}{@{\extracolsep{\fill}}cccc@{}}
        \toprule
        $k$ & BP & FA & DFA \\
        \midrule
        \multicolumn{4}{@{}l}{\emph{ResMLP $d{=}256, L{=}4$}} \\
        $0$ & $0.389 \pm 0.001$ & $0.355 \pm 0.003$ & $0.349 \pm 0.003$ \\
        $1$ & $0.565 \pm 0.003$ & $0.382 \pm 0.008$ & $0.244 \pm 0.015$ \\
        $2$ & $0.598 \pm 0.003$ & $0.349 \pm 0.016$ & $0.286 \pm 0.013$ \\
        $3$ & $0.608 \pm 0.001$ & $0.398 \pm 0.008$ & $0.296 \pm 0.008$ \\
        $4$ & $0.617 \pm 0.002$ & $0.402 \pm 0.009$ & $0.301 \pm 0.006$ \\
        \addlinespace[2pt]
        \multicolumn{4}{@{}l}{\emph{ResMLP $d{=}512, L{=}2$}} \\
        $0$ & $0.386 \pm 0.003$ & $0.359 \pm 0.000$ & $0.349 \pm 0.005$ \\
        $1$ & $0.583 \pm 0.002$ & $0.412 \pm 0.004$ & $0.226 \pm 0.015$ \\
        $2$ & $0.603 \pm 0.001$ & $0.361 \pm 0.003$ & $0.302 \pm 0.005$ \\
        \bottomrule
    \end{tabular*}
\end{table}

BP rises from $0.389$ to $0.617$ at $d{=}256$ ($+23$ points) and from $0.386$ to $0.603$ at $d{=}512$ ($+22$ points), so each trainable block adds accuracy and the Diagnostic 3 precondition holds. DFA's frozen rung is its maximum; every trained-block configuration lands below it, and the full audit sits $4.8$ and $4.7$ points under the frozen baseline at the two widths. Training deep DFA blocks lowers accuracy. The depth-utility failure spans the whole ladder. FA is intermediate, reaching $0.402$ at $d{=}256$ ($+4.7$ over frozen) and $0.361 \approx 0.359$ at $d{=}512$, where the FA depth-utility failure reproduces. FA's curve is non-monotonic, dipping at intermediate $k$.

The frozen baseline sits well above chance because the frozen stack is a structured feature map. At initialization each block contributes $\|f_l(h_l)\|/\|h_l\| \approx 0.10$, and the full four-block stack deviates from its input by $\|h_L-h_0\|/\|h_0\| = 0.196 \pm 0.003$ with $\cos(h_L, h_0) = 0.981 \pm 0.001$ (three seeds, CIFAR-10). The map is near-norm-preserving. So $k{=}0$ measures a trained embedding and readout over this fixed random map, an effectively trainable near-linear classifier on pixels.

The endpoints reproduce the external anchors: BP $k{=}L$ is $0.617$ against the audit's $0.615$, DFA $0.301$ against $0.306$, FA $0.402$ against $0.401$, and the frozen rung is $\approx 0.349$. The harness is \texttt{experiments/depth\_utility\_ladder.py}. A parameter-matched shallow baseline is not run; since BP already beats the frozen baseline by $22$--$23$ points, the depth precondition is secure.

%% file: appendice/threshold-sensitivity.tex
\section{Threshold sensitivity}
\label{app:threshold}

At the protocol thresholds, the two measurement diagnostics separate the audited methods cleanly (Table~\ref{tab:threshold}). Diagnostic 1 flags maximum per-block growth above $50$. Across three seeds BP holds at $1.0\times$ and FA at $8$--$17\times$; DFA, State Bridge, and Credit Bridge run from $694\times$ to $24{,}100\times$. Any cutoff in $[17, 694]$ gives the same verdicts. Diagnostic 2 flags $\|g_L\|$ below $10^{-7}$. BP and FA stay at $3\times10^{-4}$ and near $10^{-6}$; the three flagged methods fall to $4\times10^{-9}$ or below. Any cutoff in $[4\times10^{-9}, 9\times10^{-7}]$ is verdict-invariant.

\begin{table}[h]
    \centering
    \small
    \setlength{\tabcolsep}{6pt}
    \caption{Threshold sensitivity on the primary audit (ResMLP $d{=}256, L{=}4$, CIFAR-10, three seeds). Per-method 3-seed range of the Diagnostic 1 growth $\rho$ and the Diagnostic 2 reference norm $\|g_L\|$, with the verdict at the protocol thresholds ($\rho>50$, $\|g_L\|<10^{-7}$). \textcolor{green!50!black}{\checkmark} passes, \textcolor{red!70!black}{\ding{55}} flags.}
    \label{tab:threshold}
    \begin{tabular*}{\linewidth}{@{\extracolsep{\fill}}lcccc@{}}
        \toprule
        Method & growth $\rho$ & D1 & $\|g_L\|$ & D2 \\
        \midrule
        BP            & $1.0\times$                  & \textcolor{green!50!black}{\checkmark} & $3.1$--$4.0\times10^{-4}$ & \textcolor{green!50!black}{\checkmark} \\
        FA            & $8$--$17\times$              & \textcolor{green!50!black}{\checkmark} & $0.9$--$1.6\times10^{-6}$ & \textcolor{green!50!black}{\checkmark} \\
        DFA           & $978$--$2545\times$          & \textcolor{red!70!black}{\ding{55}} & $1.9$--$4.2\times10^{-9}$ & \textcolor{red!70!black}{\ding{55}} \\
        State Bridge  & $10{,}500$--$24{,}100\times$ & \textcolor{red!70!black}{\ding{55}} & $1.8$--$2.4\times10^{-9}$ & \textcolor{red!70!black}{\ding{55}} \\
        Credit Bridge & $694$--$1820\times$          & \textcolor{red!70!black}{\ding{55}} & $0.9$--$4.2\times10^{-9}$ & \textcolor{red!70!black}{\ding{55}} \\
        \bottomrule
    \end{tabular*}
\end{table}

Diagnostic 3 is narrower, a conservative reporting aid rather than a calibrated constant. Its $2$-point gap separates BP, which exceeds the frozen baseline by $+27$ points on the primary audit, from the failing runs. The margin is small on some settings: FA's gap is $-0.4$ points on the $d{=}512, L{=}2$ ResMLP, within the frozen baseline's seed variance of $\pm0.2$ points. The depth-utility verdict there rests on a small margin. We recommend reporting the raw gap alongside the pass/fail bit.

%% file: appendice/pitfalls.tex
\section{Pipeline pitfalls}
\label{app:pitfalls}

Seven measurement pitfalls produce a misleading verdict under the standard reporting pair. Each maps to one protocol check.

\begin{enumerate}[topsep=2pt, itemsep=3pt, leftmargin=*]
    \item \textbf{Layer-0 dominance under global averaging.} A single aggregate cosine reads as mildly positive when only the shallowest block aligns, because that block holds most of the norm budget. Report per-layer cosine, and read any aggregate only after locating where the signal sits.
    \item \textbf{Cosine against a floored reference.} When the deepest BP gradient norm has collapsed, the cosine to it measures the implementation's clamp. Record $\|g_L\|$ (Diagnostic 2) before interpreting any deep alignment number.
    \item \textbf{Batch mismatch between reference and candidate.} Different minibatches, augmentations, or dropout masks for the BP and FA vectors inflate or destabilize the reported cosine. Compute both on the same frozen forward pass when the claim is directional agreement.
    \item \textbf{Baseline mismatch for depth utility.} Comparing a partly trainable model only to full BP, or to an unmatched random baseline, overstates a weak method. Diagnostic 3 uses an architecture-matched frozen-blocks control.
    \item \textbf{Train/eval mode inconsistency.} Mode mismatches change residual scale and normalization, and therefore the diagnostic values. Fix and log the model mode for every measurement.
    \item \textbf{Post-hoc normalization that hides scale.} Renormalizing hidden states or gradients before logging erases the activation growth that Diagnostic 1 detects. Keep raw norms separate from any normalization used for plotting.
    \item \textbf{Missing null controls for interventions.} A rescue can raise cosine or accuracy for generic reasons. Pair it with a fresh-$B$ null and a matched BP control before crediting the intervention.
\end{enumerate}